\documentclass[10pt,twocolumn,letterpaper]{article}

\usepackage[pagenumbers]{cvpr} % To force page numbers, e.g. for an arXiv version
\usepackage{enumitem}
\usepackage{multirow}
\usepackage{makecell}
\usepackage{amsmath} 
\usepackage{bm}

\newcommand{\ours}{GSemSplat}
\newcommand{\ourss}{GSemSplat~}
\newcommand{\ieno}{\textit{i}.\textit{e}.}
\newcommand{\egno}{\textit{e}.\textit{g}.}

\definecolor{ForestGreen}{RGB}{0, 179, 45}

\definecolor{cvprblue}{rgb}{0.21,0.49,0.74}
\usepackage[pagebackref,breaklinks,colorlinks,allcolors=cvprblue]{hyperref}

\def\paperID{*****} % *** Enter the Paper ID here
\def\confName{CVPR}
\def\confYear{2025}

\title{\ours: Generalizable Semantic 3D Gaussian Splatting from Uncalibrated Image Pairs}

\author{
Xingrui Wang$^{1}$\thanks{This work was done when Xingrui and Hanxin were interns at Microsoft Research Asia. This paper is the result of an open-source research project starting from April 10th, 2024.} \quad Cuiling Lan$^{2}$ \quad Hanxin Zhu$^{1}$ \quad Zhibo Chen$^{1}$ \quad Yan Lu$^{2}$ \\
$^1$University of Science and Technology of China \quad $^2$Microsoft Research Asia \\
{\tt\small \{wxrui\_18264819595, hanxinzhu\}@mail.ustc.edu.cn, chenzhibo@ustc.edu.cn} \\
{\tt\small \{culan, yanlu\}@microsoft.com}
}

\begin{document}
\maketitle

\begin{abstract}

Modeling and understanding the 3D world is crucial for various applications, from augmented reality to robotic navigation. Recent advancements based on 3D Gaussian Splatting have integrated semantic information from multi-view images into Gaussian primitives. However, these methods typically require costly per-scene optimization from dense calibrated images, limiting their practicality. In this paper, we consider the new task of generalizable 3D semantic field modeling from sparse, uncalibrated image pairs. Building upon the Splatt3R architecture, we introduce \ours, a framework that learns open-vocabulary semantic representations linked to 3D Gaussians without the need for per-scene optimization, dense image collections or calibration.
To ensure effective and reliable learning of semantic features in 3D space, we employ a dual-feature approach that leverages both region-specific and context-aware semantic features as supervision in the 2D space. This allows us to capitalize on their complementary strengths. Experimental results on the ScanNet++ dataset demonstrate the effectiveness and superiority of our approach compared to the traditional scene-specific method. We hope our work will inspire more research into generalizable 3D understanding.
\end{abstract}

\section{Introduction}
\label{sec:intro}

\begin{figure}[t]
\centering
\includegraphics[width=\linewidth]{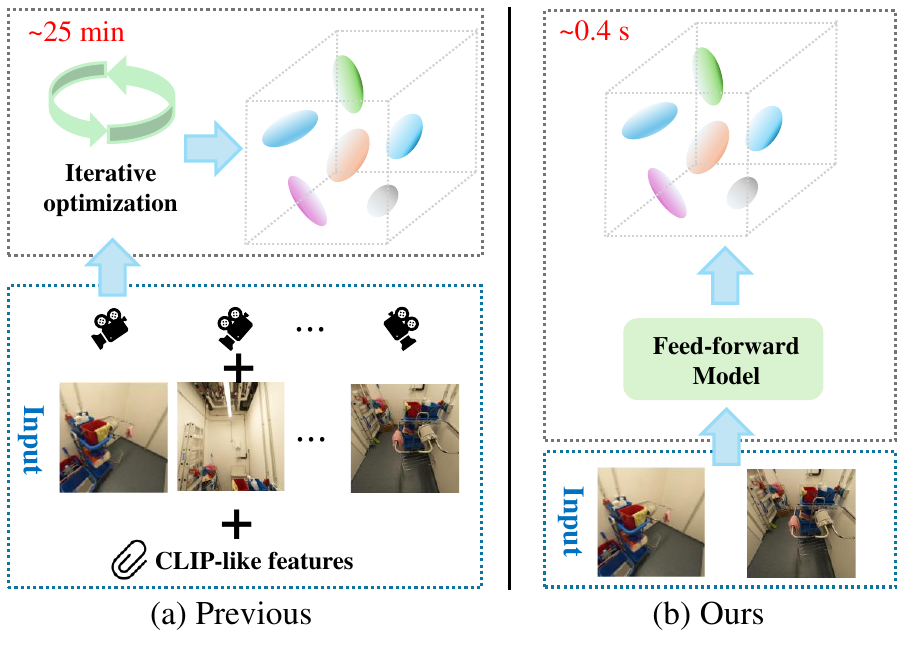} \\
\vspace{-1mm}
\caption{Comparison of methods for obtaining semantic 3D Gaussians. (a) Previous per-scene optimization-based methods require dense calibrated images and costly iterative optimization. (b) Our generalizable method allows fast feed-forward inference with spare uncalibrated (\ieno, pose-free) images as input.}
\vspace{-1mm}
\label{fig:idea}
\end{figure}

We live in and interact with the 3D world, making the 3D world modeling and understanding important. While 3D Gaussian Splatting \cite{kerbl20233d} provides explicit geometric and appearance details of a 3D scene, there is a growing demand to encapsulate semantic information into the 3D representation. Incorporating semantic information into 3D models is critical for scene understanding and human-computer interaction, enabling language-based querying and manipulation of 3D environments \cite{shen2023distilled, zhou2024feature, qin2024langsplat, qiu2024feature, xu2024tiger}. It favors many applications such as 3D semantic understanding \cite{qin2024langsplat, shi2024language}, editing \cite{xu2024tiger,qiu2024feature}, augmented/virtual reality \cite{qin2024langsplat}, robotic navigation \cite{shafiullah2022clip} and manipulation \cite{shen2023distilled}. 

Several recent works have attempted to model a 3D semantic/language field, which aims to integrate semantic information from multi-view images into Gaussian primitives for open-vocabulary querying tasks \cite{qin2024langsplat,bhalgat2024n2f2,zhou2024feature,qiu2024feature,ji2024fastlgs,hu2024semantic,shi2024language,xu2024tiger,qu2024goi}.
However, these techniques mainly focus on modeling the semantic/language field for a \emph{single scene}, through \emph{scene-specific optimization} from \emph{dense} multi-view images \cite{qin2024langsplat,bhalgat2024n2f2,zhou2024feature,qiu2024feature,ji2024fastlgs,hu2024semantic,shi2024language,xu2024tiger,qu2024goi} as illustrated in Fig.~\ref{fig:idea}~(a). Such frameworks inherit the limitations of single-scene-specific Gaussian Splatting. \emph{First}, scene-specific optimization is required for a given new scene, which is complicated and costly. \emph{Second}, camera poses are required to obtain calibrated images for optimization, adding extra effort and costs. \emph{Third}, a dense collection of dozens or hundreds of images is desired to produce high-quality results, whereas it is usually tedious and inconvenient. These hinder the easy use of such frameworks for modeling and understanding 3D scenes. Consequently, there is a pressing need for methods that can generalize across scenes without requiring extensive per-scene optimization and densely calibrated images.

In this paper, we consider the new problem of generalizable 3D semantic field modeling from sparse, uncalibrated images, \ieno, learning generalizable, pose-free semantic 3D Gaussian Splatting from uncalibrated sparse images, without requiring per-scene optimization. This problem is rarely studied yet. Fig.~\ref{fig:idea} illustrates the differences between our method and most previous methods \cite{qin2024langsplat, zhou2024feature, qiu2024feature}. Our approach is inspired by the recent advances in generalizable 3D Gaussian Splatting~\cite{Fu_2024_CVPR, fan2024instantsplat, schmidt2024look,fei2024pixelgaussian,zhang2024transplat}, particularly the Splatt3R network \cite{smart2024splatt3r}. We build upon Splatt3R architecture to create a framework, \ours, for generalizable open-vocabulary scene understanding. 
Specifically, we aim to learn semantic representations bonded to 3D Gaussians from abundant scenes without requiring human annotations of semantics. Similar to the observation in scene reconstruction \cite{mildenhall2021nerf, niemeyer2022regnerf, jain2021putting, seo2023mixnerf, seo2023flipnerf, zhu2024vanilla, smart2024splatt3r}, the traditional per-scene optimization of semantics from sparse images faces failures due to insufficient information. By learning priors across abundant scenes, our \ourss can effectively capture semantics from sparse input images.

\begin{figure}[t]
\centering
\includegraphics[width=\linewidth]{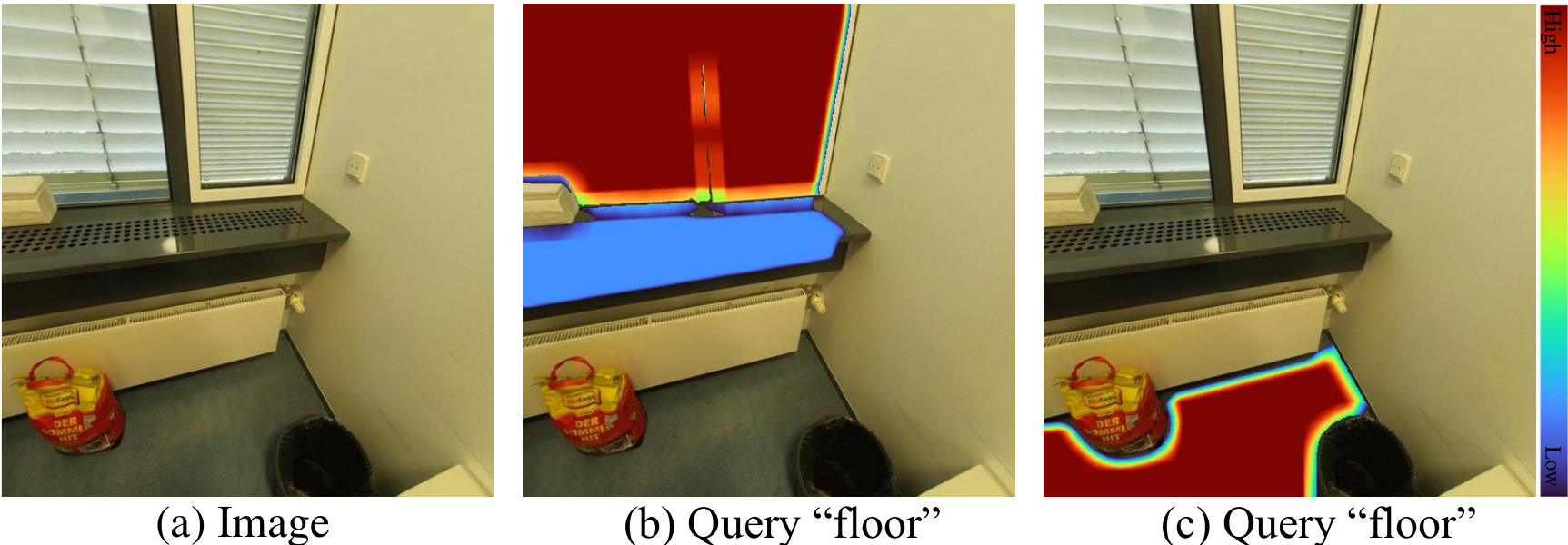} \\
\caption{Given an input image (a), we present the response map when we use the text query of ``floor" to retrieve the correlated content from the region-specific feature in (b), and from the context-aware feature in (c). For the region-specific feature on the floor region, due to lack of context, the region presents a low correlation with the query. In contrast, the floor region presents a high correlation for context-aware features as shown in (c).}
\vspace{-2mm}
\label{fig:SAMorPatch-Patch}
\end{figure}

Most previous works distill and lift the 2D language-aligned vision CLIP features \cite{radford2021learning} or variants \cite{oquab2023dinov2} to 3D space for scene-specific optimization. Some works extract a hierarchy of CLIP features from cropped image patches as the semantic representations \cite{liu2023weakly, kerr2023lerf}. Such representations are patchy and suffer from unclear object boundaries \cite{liu2023weakly}. LangSplat \cite{qin2024langsplat} addresses boundary ambiguity by generating segmentation masks by SAM \cite{kirillov2023segment} and extracting CLIP features for each masked region, which we refer to as region-specific CLIP features. Region-specific CLIP features exhibit excellent object-distinguishing capability with clear boundaries but suffer from the absence of context (around the object). For some masked regions, the extracted CLIP features are unreliable and incorrect. For example, the ``floor" region only (without neighboring context) could be mistakenly recognized as ``ceiling" even by humans and does not obtain a high correlation score as shown in Fig.~\ref{fig:SAMorPatch-Patch}. MaskCLIP \cite{dong2023maskclip} directly generates dense semantic features. This preserves context information but suffers from feature inaccuracy due to the inevitable interference from the background or neighboring objects over the current pixels. How to construct and explore reliable features to distill semantics is still under-explored. We propose to leverage both region-specific CLIP features and context-aware CLIP features for each Gaussian for capturing complementary information.

\begin{figure*}[t]
\centering
\includegraphics[width=\linewidth]{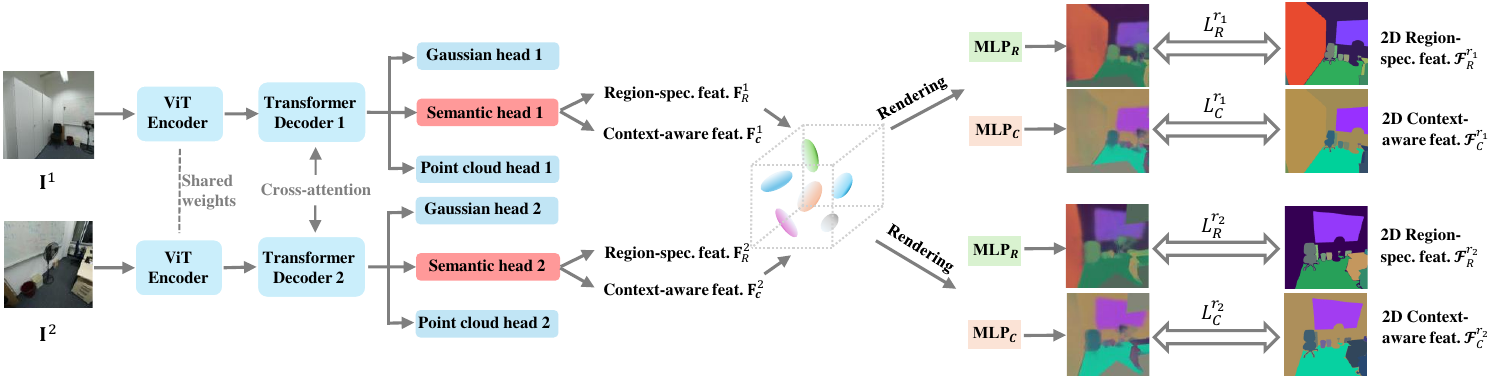} \\
\vspace{-1mm}
\caption{Illustration of our overall framework \ours, for generalizable open-vocabulary 3D scene understanding, obviating the need for costly and complicated per-scene optimization, tedious extensive image collection and calibration processes. We use the generalizable 3D Gaussian Splatting architecture, Splatt3R~\cite{smart2024splatt3r}, as our base network, which predicts 3D Gaussians from uncalibrated image pairs. We introduce a new semantic head that predicts the low-dimensional semantics associated with each Gaussian. Without relying on costly human annotation of 3D semantics, we distill the 3D semantic information from the 2D semantic features. Particularly, \ourss simultaneously predicts region-specific semantic features, and context-aware semantic features, facilitating text querying to localize the more reliable semantics. The predicted low-dimensional semantic features are transformed into high-dimensional counterparts via MLP blocks for open-vocabulary semantic understanding.}
\vspace{-3mm}
\label{fig:framework}
\end{figure*}

We illustrate our overall framework in Fig.~\ref{fig:framework}. We use the generalizable 3D Gaussian Splatting architecture Splatt3R as our base network, which predicts 3D Gaussians from uncalibrated image pairs. 
We introduce a new semantic head designed to infer detailed semantic information for each Gaussian. We distill the 3D semantic information from two sets of 2D semantic features: region-specific CLIP features and context-aware CLIP features, to enable more reliable semantic representations. This dual-feature strategy enhances the accuracy of text-based querying for identifying optimal matches. The semantic head outputs a low-dimensional region-specific feature and a low-dimensional context-aware feature bound to each predicted Gaussian, where the low-dimensional features can be projected to high-dimensional features through MLP blocks to facilitate text querying. 

In summary, our main contributions are threefold:
\begin{itemize} [noitemsep,nolistsep,leftmargin=*]
    \item We propose a new framework, \ours, for generalizable 3D semantic field modeling from sparse, uncalibrated (pose-free) images. To the best of our knowledge, this work is the first to build generalizable 3D semantic Gaussians from uncalibrated sparse images. 
    \item We leverage Splatt3R as our base architecture to empower the semantic prediction capability. We construct dual features, \ieno, region-specific CLIP features, and context-aware CLIP features,  with complementary merits, for effective distillation of 3D semantics.
    \item Experimental results demonstrate the effectiveness of our framework. Our model achieves superior performance with 4000 $\times$ faster running speed compared with the per-scene optimization-based method. 
\end{itemize}

We build a benchmark based on the ScanNet++ dataset \cite{yeshwanth2023scannet++} for studying this new problem. We hope our work will inspire more investigations on this important and practical task to advance generalizable 3D understanding.

\section{Related Work}
\label{sec:relatedwork}

\noindent\textbf{3D Scene Modeling} 3D scene modeling has long been a pivotal area of research in computer vision, graphics, and robotics. In recent years, neural representations for 3D modeling, \egno, Neural Radiance Fields (NeRFs) \cite{mildenhall2021nerf}, have demonstrated remarkable results in synthesizing novel views. NeRFs represent 3D scenes by fully connected neural networks, where volume density and radiance can be queried by input position and view direction vectors. However, NeRFs suffer from slow training and rendering speed~\cite{chen2021mvsnerf, wang2021ibrnet}. Recent 3D Gaussian Splatting (3DGS) explicitly models the 3D scene as a collection of Gaussians, representing the geometric and appearance details \cite{kerbl20233d}. 3DGS sets a new standard on both speed and rendering quality, by using a fast tile-based rasterization.  

These methods \cite{kerbl20233d,yu2024mip,lu2024scaffold,cheng2024gaussianpro,niemeyer2024radsplat} in general require dozens or hundreds of calibrated images (with known camera poses) to produce high-quality results, through expensive, iterative, per-scene optimization, which are unable to utilize cross-scene priors. To mitigate these problems, generalizable 3D reconstructors are designed to predict pixel-aligned features from sparse calibrated images using feed-forward networks \cite{charatan2024pixelsplat,chen2024mvsplat,szymanowicz2024splatter,szymanowicz2024flash3d,wewer2024latentsplat,wang2024freesplat,tang2025lgm,zhang2025gs,xu2024grm}. These methods still require known camera parameters, which hinders the usage of uncalibrated input images. Splatt3R \cite{smart2024splatt3r} addresses this challenge by unifying the Gaussian predictions into entities under the same reference view, despite some Gaussians being predicted from other reference views. Motivated by Splatt3R \cite{smart2024splatt3r}, we aim to build a framework that is capable of modeling generalizable 3D \emph{semantic field} from sparse uncalibrated images, for effective open-vocabulary semantic understanding.

\noindent\textbf{3D Semantic/Language Field} Considering the fast speed and rendering quality of Gaussian Splatting, many recent works \cite{qin2024langsplat,bhalgat2024n2f2,zhou2024feature,qiu2024feature,ji2024fastlgs,hu2024semantic,shi2024language,xu2024tiger,qu2024goi} attempted to incorporate semantic information to Gaussians to obtain semantic/language fields, which facilitates scene understanding and manipulation. However, all these methods rely on single-scene-specific optimization, being complicated and costly. In practical applications, feed-forward inference without re-optimization from sparse uncalibrated images is highly desired, avoiding costly per-scene optimization, tedious collection of dense views, and estimation of camera poses. To the best of our knowledge, there are very few works to support this yet. Our main contribution lies in that we design a framework that enables generalizable 3D semantic field learning, being application-friendly. 

Many works elevate the 2D semantic features from vision-language aligned model CLIP into 3D space as semantic information~\cite{qin2024langsplat,bhalgat2024n2f2,qiu2024feature,ji2024fastlgs,shi2024language,xu2024tiger,qu2024goi}, where text query is used to identify relevant semantic regions. Considering CLIP is originally designed to extract image-level features, some works extract a hierarchy of CLIP features from cropped image patches for pixel-level feature extraction by feeding the cropped images into CLIP vision encoder \cite{liu2023weakly, kerr2023lerf}. The resulting relevancy maps are usually patchy and blurry. LangSplat \cite{qin2024langsplat} addresses the patchy and blurry effects by using SAM \cite{kirillov2023segment} to semantically segment the regions and feed each region into CLIP to extract the region feature. While preserving high boundary clarity, some objects/stuff, suffer from semantic ambiguity due to a lack of enough context for extracting region features. For example, given a region of ``floor", it is difficult to distinguish between ``floor" and ``celling". TIGER \cite{xu2024tiger} employs MaskCLIP \cite{dong2023maskclip} to directly generate dense semantic features and then upsample these features to the pixel level. This enables the preservation of context information for the extracted features. However, the features are not accurate due to the inevitable influence of the background or neighboring objects for the current pixels. 
How to construct the features for reliable supervision of the 3D semantic features is still under-explored.  
In this paper, we propose to leverage both region-specific CLIP features and context-aware CLIP features for more reliable semantic learning.

\section{Method}
\label{sec:method}

We propose a \textbf{G}eneralizable 3D \textbf{Sem}antic Gaussian \textbf{Splat}ting framework, \ours, for efficient 3D open-vocabulary scene understanding. Sec. \ref{subsec:overview} gives an overview of \ourss. We review the base network Splatt3R \cite{smart2024splatt3r} in Sec. \ref{subsec:Splatt3R}. In Sec. \ref{subsec:method}, we elaborate on the 3D semantic learning.

\subsection{Framework Overview}
\label{subsec:overview}

In testing, for a new scene, as illustrated in Fig.~\ref{fig:framework}, given two uncalibrated input images, \ourss outputs the 3D Gaussians with each accompanied by two semantic features: one region-specific feature and one context-aware feature, through one feed-forward run. 
Then, we can query the scene for scene understanding by matching the language query with the semantic features to identify the highly correlated regions.

The network is trained with abundant scenes to learn the priors from scene reconstruction and semantic reconstruction. As shown in Fig.~\ref{fig:framework}, on top of Splatt3R \cite{smart2024splatt3r} (which is marked by blue), we introduce a new semantic head that outputs a low-dimensional region-specific feature and a low-dimensional context-aware feature as the semantic features for each pixel. The semantics are bound to the predicted 3D Gaussian of that pixel. We refer to a 3D Gaussian with semantics as \emph{semantic 3D Gaussian}. We obtain a union of the inferred semantic 3D Gaussians from the two reference views. After rendering the semantic 3D Gaussians to a target view, the semantic features are projected/de-compressed to high-dimensional region-specific CLIP features and context-aware CLIP features through an MLP subnet, respectively. These features are supervised by the extracted 2D region-specific CLIP features and context-aware CLIP features to train the semantic head.

\subsection{Review of Base Network Splatt3R}
\label{subsec:Splatt3R}

Recently, DUSt3R \cite{wang2024dust3r} and its follow-up works MASt3R \cite{leroy2024grounding}, and Splatt3R \cite{smart2024splatt3r}, obviate the need for camera poses for inferring 3D locations (point maps) \emph{by predicting the 3D locations (and Gaussians \cite{smart2024splatt3r} in Splatt3R) of all points in the first reference image's camera frame}. This avoids the need for the transformation of other views' points to the same view based on camera parameters. We take Splatt3R \cite{smart2024splatt3r} as our base network, which predicts 3D Gaussian from a pair of images with unknown camera parameters.

The Splatt3R network (as marked by blue in Fig.~\ref{fig:framework}) consists of a weight-shared ViT encoder, two transformer decoders followed by a Gaussian head, and a point cloud head. Given two input reference images $\mathbf{I}^1$, $\mathbf{I}^2 \in \mathbb{R}^{H\times W \times 3}$ from the two reference views, the Gaussian head 1 predicts the Gaussian parameters for each pixel of $\mathbf{I}^1$ at its camera view while the Gaussian head 2 predicts the Gaussian parameters for each pixel of $\mathbf{I}^2$ at $\mathbf{I}^1$'s camera view too. The point cloud head predicts the 3D point location ($\mathbf{p}$) for each pixel. The Gaussian head predicts the parameters of a Gaussian for each pixel: covariances $\mathbf{\Sigma}$ (parameterized by rotation $\mathbf{q}\in \mathbb{R}^4$ and scale $\mathbf{s} \in \mathbb{R}^3$), spherical harmonics ($\mathbf{S} \in \mathbb{R}^{3 \times d}$), opacities ($\alpha \in \mathbb{R}$), and offset ($\mathbf{\Delta}\in \mathbb{R}^3$) to 3D point location where the center of a Gaussian is $\mathbf{\mu} = \mathbf{p} + \mathbf{\Delta}$. Thus, a 3D Gaussian point can be formally defined as $G(\mathbf{x}\vert \mathbf{\mu}, \mathbf{\Sigma}) = e^{-\frac{1}{2}(\mathbf{x}-\mathbf{\mu})^T \mathbf{\Sigma}^{-1} (\mathbf{x}-\mathbf{\mu})}$.
The union of the predicted Gaussians for all pixels from the two input images $\mathbf{I}^1$ and $\mathbf{I}^2$ constructs the final set of Gaussians for the scene. 

The Gaussian head is optimized through the reconstruction supervision over rendered novel views. The remaining subnetworks, including the ViT encoder, transformer decoders, and the point cloud head, are loaded from the pre-trained MASt3R \cite{leroy2024grounding} model and frozen during optimization. 

\subsection{Semantic Learning}
\label{subsec:method}

Our objective is to predict semantic 3D Gaussians from a pair of uncalibrated images. To this end, as illustrated in Fig.~\ref{fig:framework}, on top of Splatt3R, we add a new head (in parallel to the Gaussian head), dubbed semantic head, to predict the semantic features to the Gaussian for each pixel. 
In this way, each Gaussian is equipped with its 3D semantics.   

To avoid the use of costly human annotations of semantics for supervision, similar to previous methods \cite{qin2024langsplat, shi2024language, liao2024clip}, we elevate the 2D semantic features extracted from CLIP vision encoder to 3D space. CLIP (Contrastive Language-Image Pre-Training) \cite{radford2021learning} aligns vision features (at the image level) with language features, facilitating retrieval of vision features based on open-vocabulary query. Previous works usually leverage CLIP features or their variants for learning 3D semantics. 

There are two challenges to address. 1) How to construct 2D semantic features for more reliable supervision? We propose to jointly learn region-specific features and context-aware features to exploit their respective merits. As illustrated in Fig.~\ref{fig:framework}, the semantic head predicts region-specific features and context-aware features by a two-branch design. 2) The high dimension of semantic features would largely increase the memory and computational costs. How to reduce the cost? We learn low-dimension semantic features and project them to high-dimension after the rendering for 2D semantic supervision. We elaborate on the design below. 

\noindent\textbf{2D Semantic Features for Supervision} Recently, LangSplat \cite{qin2024langsplat} uses the Segment Anything Model (SAM \cite{kirillov2023segment}) to semantically segment an image into regions and feed each masked region into CLIP to extract the region feature. We refer to such features as region-specific features (or SAM-isolated CLIP features). \emph{This strategy avoids the inferences from neighboring regions on the feature extraction.} It also obviates the blurry and patchy effects. However, due to the lack of context information for feature extraction, some region features present ambiguity and wrong semantics. As an example shown in Fig.~\ref{fig:SAMorPatch-Patch}, when we query with text ``floor", the floor region in the region-specific feature map presents low responses while the windowsill region presents higher responses. 

As a remedy, we propose to learn both region-specific features and context-aware features using two branches of the semantic head. For the context-aware feature, we use the SAM-enhanced CLIP feature as the 2D supervision on the rendered view. Raw CLIP feature map extracted from an image contains context information. But it is noisy and low-resolution. Similar to Feature Splatting \cite{qiu2024feature}, we generate high-quality feature maps by pooling within region masks produced by SAM and filling the pooled feature embeddings to high-resolution feature maps. Fig.~\ref{fig:SAMorPatch-Patch} (c) shows an example of retrieving ``floor" over the context-aware feature map, where the floor region can be correctly identified.

\begin{figure}[t]
\centering
\includegraphics[width=\linewidth]{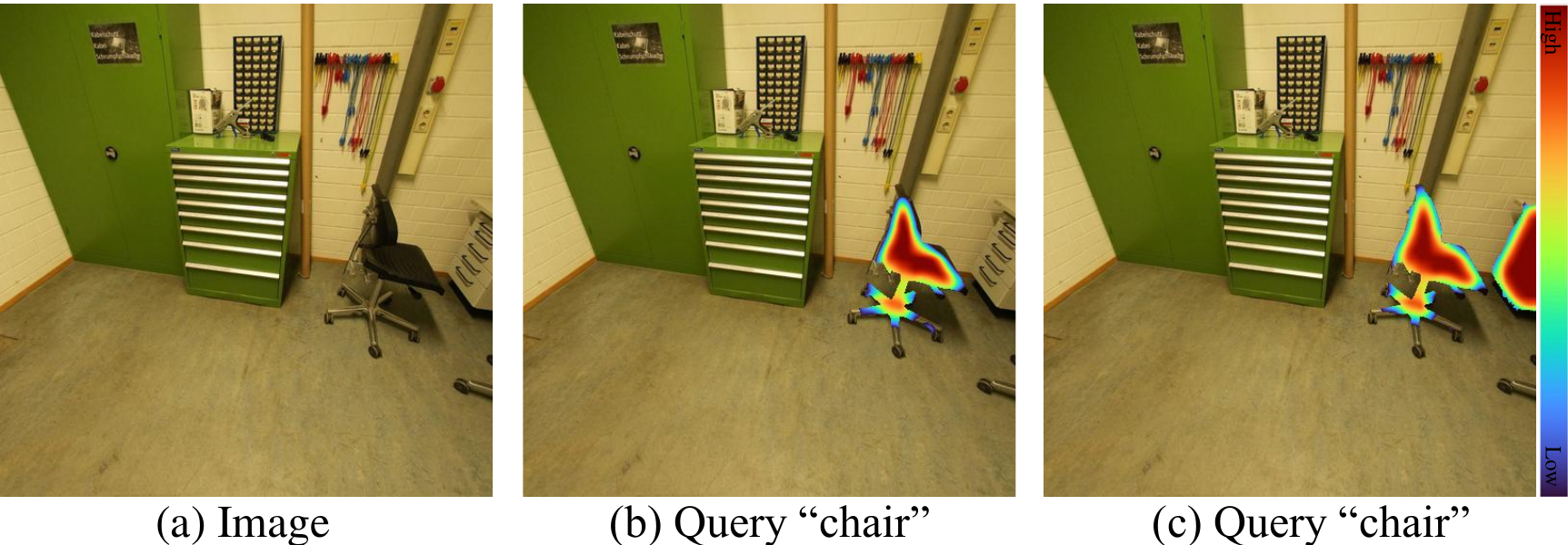} \\
\caption{Given an input image (a), we present the response map when we use the text query of ``chair" to retrieve the correlated content from the region-specific feature in (b), and from the context-aware feature in (c). For the context-aware feature, the cabinet region mistakenly presents a high correlation with the query due to interference from the neighboring chair region. In contrast, the cabinet region presents a low correlation for the region-specific feature as shown in (b).}
\vspace{-1mm}
\label{fig:SAMorPatch-SAM}
\end{figure}

We conducted experiments to analyze the necessity of exploiting both the region-specific features and context-aware features on the extracted 2D semantic features. Here, we treat each query over an image as a sample. Experimental results show that 33.72\% samples prefer region-specific features while 66.28\% samples prefer context-aware features, among the samples that their IoU (Intersection over Union) difference is larger than 5\%. Please refer to Supplementary for more details. In general, we found that context-aware features yield better results when the context is necessary to identify an object or stuff (such as the floor region in Fig.~\ref{fig:SAMorPatch-Patch}). For objects/stuff where the segmented region contains sufficient details to recognize the category  (such as the chair in Fig.~\ref{fig:SAMorPatch-SAM}), region-specific features yield better results, with more accurate semantic feature representations that are less affected by neighboring regions.

\noindent\textbf{Semantic Head Prediction and Optimization} As illustrated in Fig.~\ref{fig:framework}, we add a semantic head 1 for reference view 1 branch and a semantic head 2 for reference view 2 branch. For the $i^{th}$ input view (where $i=1,2$), the semantic head takes the set of decoder tokens $T^i$ as input and outputs the compressed region-specific feature map $\textbf{F}_R^{i} \in \mathbb{R}^{H\times W\times K}$ and the compressed context-aware feature map $\textbf{F}_C^{i} \in \mathbb{R}^{H\times W\times K}$:
\begin{equation}
    \textbf{F}_R^{i}, \textbf{F}_C^{i} = SemHead^{i}(T^i).
    \label{eq:semantichead}
\end{equation}
Here, to reduce the memory and computational costs, we regress the semantic feature at a low dimensional space, with the feature dimension $K \ll D$, where $D=512$ denotes the original dimension of the 2D semantic features. We set $K$ as 16 by default.  For the $i^{th}$ view, the semantic 3D Gaussian point for pixel $(m,n)$ can then be represented as $\{G(\mathbf{x} \vert \mathbf{\mu}_{m,n}^i, \mathbf{\Sigma}_{m,n}^i), \mathbf{F}_R^i(m,n),\mathbf{F}_C^i(m,n)\}$.

In training, we render the low-dimensional features to two target views (\ieno, view $r_1$ and $r_2$). Two MLP subnets $MLP_R$ and $MLP_C$ are then used to project the low-dimensional region-specific features and context-aware features to high-dimensional features, supervised by the high-dimensional 2D region-specific features and context-aware features, respectively.

For the $i^{th}$ target/rendering view $r_i$, we render its low-dimensional region-specific features and context-aware features to this view followed by the projection through the MLP subnets. We obtain the reconstructed high-dimensional region-specific features $\bar{\pmb{\mathcal{F}}}_R^{r_i}$ and context-aware features $\bar{\pmb{\mathcal{F}}}_C^{r_i}$. For region-specific features, the regression loss for a pixel at  position $(m,n)$ of rendering view $r_i$ is as 
\begin{equation} 
     \mathcal{L}_R^{r_i} (m,n) = 1-\cos\left( {\bar{\pmb{\mathcal{F}}}_R^{r_i}(m,n), \pmb{\mathcal{F}}_R^{r_i}(m,n)} \right),  
\label{eq:region_loss}
\end{equation}
where $\pmb{\mathcal{F}}_R^{r_i}(m,n)$ denotes the region-specific features used for supervision, $\cos(\cdot,\cdot)$ denotes cosine similarity. Similarly, we denote the regression loss for context-aware features as $ \mathcal{L}_C^{r_i} (m,n)$. 
The final training objective is as
\begin{equation}
\begin{aligned}  
\mathcal{L}_{sem} & = \sum_{i \in \{1,2\}} \sum_{m=1}^H  \sum_{n=1}^W \mathcal{L}_R^{r_i} (m,n) + \mathcal{L}_C^{r_i} (m,n).
\end{aligned}
\label{eq:confidence_aware_loss}
\end{equation}

\noindent\textbf{Open-vocabulary Querying} Following LERF \cite{kerr2023lerf} and LangSplat \cite{qin2024langsplat}, we compute the regularized relevancy score for the text query. We follow the strategy used in LERF \cite{kerr2023lerf} to choose the region-specific or context-aware that yields a higher regularized relevancy score for each query. Our statistical analysis shows that this strategy can hit better features at an accuracy of 70\% when we do the study on the extracted 2D semantic features. Please see our Supplementary for more details. For the 3D object localization task, we directly choose the point with the highest score. For the 3D semantic segmentation task, we filter out points with scores lower than a chosen threshold as the predicted mask region. 

\begin{table*}[t]
\centering
\normalsize
\setlength{\tabcolsep}{0.45mm}
\resizebox{1\textwidth}{!}{ 
\begin{tabular}{c|cc|cc|cc|cc|cc|cc|cc|cc}
\toprule
\multirow{2}{*}{Method} 
                        & \multicolumn{2}{c|}{office} & \multicolumn{2}{c|}{classroom-1} & \multicolumn{2}{c|}{utility room} & \multicolumn{2}{c|}{workshop}  & \multicolumn{2}{c|}{printing room} & \multicolumn{2}{c|}{kitchen} & \multicolumn{2}{c|}{meeting room} & \multicolumn{2}{c}{classroom-2} \\ \cmidrule{2-17}
                        & mIoU$\uparrow$ & LA$\uparrow$ & mIoU$\uparrow$   & LA$\uparrow$ & mIoU$\uparrow$ & LA$\uparrow$ & mIoU$\uparrow$ & LA$\uparrow$ & mIoU$\uparrow$ & LA$\uparrow$ & mIoU$\uparrow$   & LA$\uparrow$ & mIoU$\uparrow$ & LA$\uparrow$ & mIoU$\uparrow$ & LA$\uparrow$\\ \midrule
LangSplat \cite{qin2024langsplat}               & 57.7 & 83.3  & 57.5 & 66.7  & 41.3 & 100  & 68.8 & 75.0   & 31.4 & 40.0  & 64.0 & 66.7  & 41.3 & 66.7  & 53.5 & 83.3 \\ \midrule
\ours~(Ours)                  & \textbf{86.1} & \textbf{100} & \textbf{78.9} & \textbf{100} & \textbf{78.9} & {100} & \textbf{76.0} & \textbf{100} & \textbf{53.4} & \textbf{80.0} & \textbf{84.1} & \textbf{100} & \textbf{73.6} & \textbf{100} & \textbf{73.8} & \textbf{100} \\

\bottomrule
\end{tabular}
}
\caption{Performance comparison between LangSplat and our \ourss on segmentation measured by mIoU score (\%), and on localization measured by accuracy (LA)(\%).}
\label{tab:Comparison Results}
\end{table*}

\section{Experiments}
\label{sec:experiments}
\subsection{Settings}
\label{sec:settings}

\noindent\textbf{Evaluation Metrics} We evaluate semantic segmentation performance using mean Intersection-over-Union (mIoU) and Localization Accuracy (LA)~\cite{qin2024langsplat}. Following Splatt3R~\cite{smart2024splatt3r}, all metrics are computed after applying loss masks to the rendered semantic images to exclude the calculation on invisible regions. We assess localization accuracy by checking if the point with the highest response falls within the corresponding ground truth semantic mask.

\noindent\textbf{Training Data} We use the ScanNet++ dataset\cite{yeshwanth2023scannet++}, which contains high-fidelity 3D geometry and high-resolution RGB images of indoor scenes, for training and testing. Following the setting of Splatt3R~\cite{smart2024splatt3r} that was originally set for 3D reconstruction, we use the official training (230 scenes) and validation (50 scenes) splits, excluding frames marked as ``bad'' and scenes with frames lacking valid depth. All frames are cropped and resized to a resolution of 512$\times$512. 

\noindent\textbf{Testing Data} The ScanNet++ validation split provides detailed semantic and instance annotations of 3D meshes. We project the 3D semantic labels of a scene onto 2D images, followed by distortion correction and resizing to generate ground-truth semantic masks for evaluation. We focus on commonly used categories, such as ``ceiling", ``floor",  ``wall", and ``chair", selected from the top 100 semantic classes in ScanNet++, as the queries. The queries for each scene vary according to scene-specific characteristics. Please refer to our Supplementary for more details. During testing, we render the semantic features from the predicted 3D semantic field to the target view for the open-vocabulary text querying. Given that Splatt3R provides the testing subsets, we evaluate the subset with at least 50$\%$ direct pixel correspondence. More results on other ratios of pixel correspondence can be found in Supplementary.

Moreover, we also tested our model on in-the-wild testing examples to evaluate the generalization capability. 

\noindent\textbf{Implementation Details} For our scheme, during training, we input two context images and render two target views for supervision on semantics. We extract 2D region-specific features and context-aware features for target view images as supervision. We choose views such that each context-context pair and each target-context pair both share a minimum of 30$\%$ direct pixel correspondence.
Consistent with Splatt3R, we also apply loss masks to mitigate the impact of regions invisible in the context images.

\noindent\textbf{Baseline Method for Comparison}
To the best of our knowledge, there is no prior work that addresses generalizable 3D semantic field modeling from sparse pose-free images. As an alternative, we compare our method to the per-scene optimization-based method LangSplat~\cite{qin2024langsplat}. 

\begin{figure}[t]
\centering
\includegraphics[width=\linewidth]{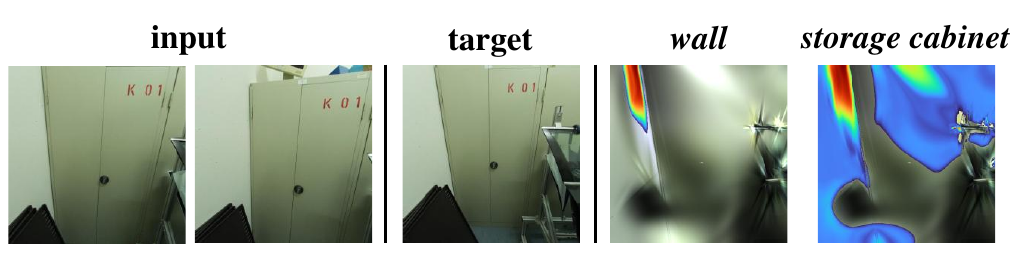} \\
\caption{Open-vocabulary semantic segmentation results (the last two columns) of LangSplat are poor when it is trained/optimized with \emph{sparse views} (two context views in the first two columns). }
\vspace{-1mm}
\label{fig:sparse view}
\end{figure}

Actually, LangSplat performs poorly in reconstructing semantic fields from sparse views. As the example shown in Fig.~\ref{fig:sparse view}, when we train LangSplat with only two context views and render the scene on the target view, the reconstructed scene appears blurred (see the fourth column) and the semantics are inaccurate (see the fourth and fifth columns). 
In the following experiments, to enable high performance of the baseline for comparison, we train LangSplat with dense views.       
Particularly, in each scene, we use all images and their semantic features (\ieno, SAM-isolated CLIP features) within the highest and lowest indices range across the two context views and the testing view to train LangSplat, with the testing view excluded. The number of images for optimizing a scene ranges from 60 to 400. 
In contrast, our scheme obtains the corresponding semantic field using only the two input context images.

\subsection{Results}

\begin{figure*}[t]
\centering
\includegraphics[width=\linewidth]{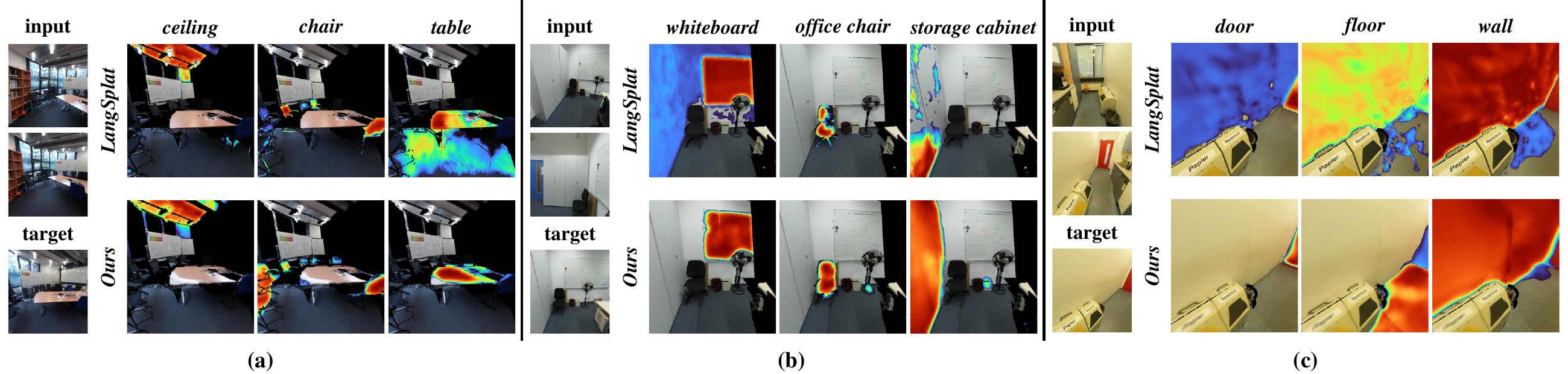} \\
\caption{Qualitative comparison between LangSplat and our \ours. We present the correlation maps between the text queries and learned semantic features on the rendered target images. Here, low correlation ($<$ 0.5) responses were filtered out for clearer presentation.}
\vspace{-1mm}
\label{fig:quantitative comparison}
\end{figure*}

\noindent\textbf{Comparison with LangSplat} The per-scene optimization process of LangSplat is complicated (which requires camera parameter estimation, scene-specific autoencoder training, and two-step training) and it is labor-intensive to test all the scenes. Therefore, we select eight representative scenes from the validation set for comparison.  For each scene, we randomly select one sample composed of two context images and one target/test image for evaluating the quality of the rendered target view semantics. Tab.~\ref{tab:Comparison Results} shows the comparison between LangSplat and our \ours. Our \ourss consistently outperforms LangSplat in both segmentation (measured by mIoU score) and localization (measured by localization accuracy). 

Fig~\ref{fig:quantitative comparison} shows visualization comparisons of the correlation maps from the given text queries for LangSplat and our model. We observe that our model achieves more accurate identification of the queried objects/stuff. In Fig~\ref{fig:quantitative comparison}~(a), ``chair" is more accurately identified by our model while some chairs are missed by LangSplat. In Fig~\ref{fig:quantitative comparison}~(b), ``storage cabinet is only partially identified by LangSplat. In addition, LangSplat mistakenly recognizes the wall as ``floor" in Fig~\ref{fig:quantitative comparison}~(c), due to lack of sufficient context. Our method achieves more precise responses with clear boundaries and demonstrates robustness across various scenes. We provide more visualization results in our Supplementary.

\begin{figure}[t]
\centering
\includegraphics[width=\linewidth]{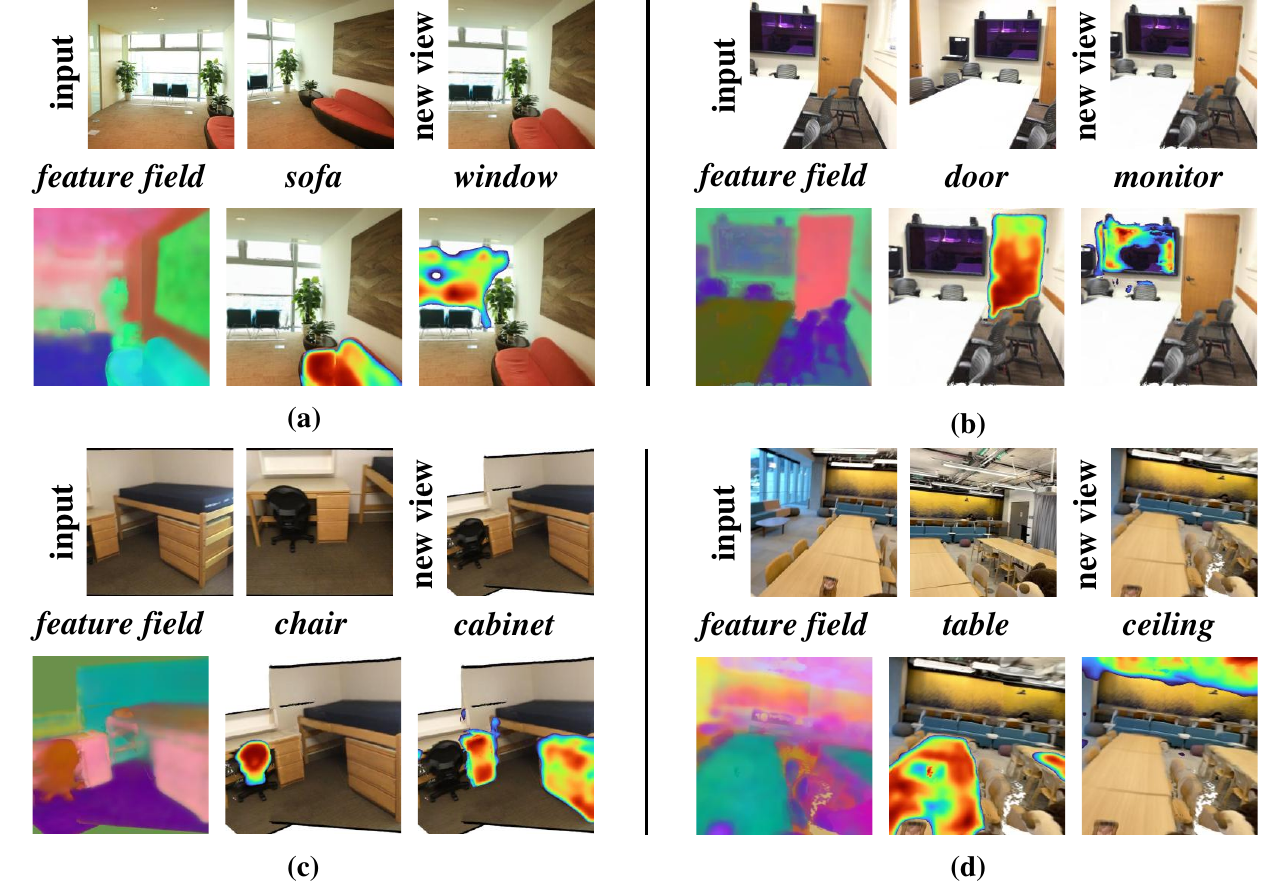} \\
\vspace{-1mm}
\caption{Zero-shot testing results of \ourss on (a) our captured images, (b-c) ScanNet \cite{dai2017scannet}, and (d) LERF \cite{kerr2023lerf}.}
\vspace{-1mm}
\label{fig:Generalization Results}
\end{figure}

\noindent\textbf{Generalization Results} To evaluate the generalization capability of our model, we conducted tests on scenes from the ScanNet \cite{dai2017scannet} and LERF~\cite{kerr2023lerf} datasets, as well as our captured real-world images. The results are presented in Fig.~\ref{fig:Generalization Results}. For visualization of the semantic field from a new view, we reduce the dimension of the rendered semantic features from 512 to 3 dimensions via PCA. We find that \ourss effectively perceives and distinguishes distinct objects/stuff, yielding accurate responses across diverse query texts.

\noindent\textbf{Runtime Comparison on New Scene} We compare the runtime for obtaining the semantic field for a given new scene. For LangSplat, we measure the time required for 30,000 iterations of optimization to reconstruct the semantic field. All evaluations are conducted on one NVIDIA A100 GPU, and the comparison results are presented in Tab.~\ref{tab:runtime comparisons}. Given a new scene, since LangSplat requires iterative optimization, it incurs a higher time cost (about 25 minutes). In contrast, our \ourss achieves very fast inference through a feed-forward manner (0.35 seconds), being 4000 $\times$ faster. 

\begin{table}[t]
    \centering
\resizebox{0.33\textwidth}{!}{     
    \begin{tabular}{c|c}
    \toprule
    Method  & Reconstruction Time$\downarrow$ \\  \midrule
    LangSplat     & 25~min~6s  \\
    \ours~(Ours)          & 0.35~s     \\
    \bottomrule
    \end{tabular}
    }
    \caption{Runtime comparison of LangSplat and \ours.}
    \label{tab:runtime comparisons}
    \vspace{-3mm}
\end{table}

\subsection{Ablation Studies}
\label{sec:ablation studies}

\noindent\textbf{Effectiveness of Dual Features} To verify the effectiveness of jointly exploiting region-specific and context-aware features, we test the performance of \ourss when only a single type of feature is used. We use the same scenes as listed in Tab.~\ref{tab:Comparison Results}. To obtain statistical results, for each scene, we randomly sample 100 samples (2 context images and 1 target image) for testing. The average results for each scene are summarized in Tab.~\ref{tab:Dual-semantic Features}. We observe that using only region-specific or context-aware features leads to a performance drop when compared with our final scheme. Combining them facilitates the use of their respective strengths to provide more reliable responses to given queries, resulting in higher localization accuracy and more precise segmentation. Please see more results under the same setting as Tab.~\ref{tab:Comparison Results} in Supplementary.

\begin{table}[t]
\centering
\small
\setlength{\tabcolsep}{1mm}
\resizebox{0.42\textwidth}{!}{ 
\begin{tabular}{c|cc|cc|cc}
\toprule
\multirow{2}{*}{Scene} & \multicolumn{2}{c|}{Region} & \multicolumn{2}{c|}{Context} & \multicolumn{2}{c}{Both} \\ \cmidrule{2-7}
                       & mIoU$\uparrow$     & LA$\uparrow$        & mIoU$\uparrow$     & LA$\uparrow$          & mIoU$\uparrow$      & LA$\uparrow$   \\  \midrule
office             & 51.0   & 77.5    & 52.4   & 81.7      & \textbf{57.8}    & \textbf{84.9}   \\  
classroom-1        & 43.7   & 68.9    & 45.5   & 77.7      & \textbf{48.2}    & \textbf{78.6}   \\ 
utility room       & 41.9   & 70.1    & 35.1   & 68.2      & \textbf{47.9}    & \textbf{76.3}     \\ 
workshop           & 42.4   & 76.8    & 38.1   & 72.1      & \textbf{48.9}    & \textbf{77.7}     \\ 
printing room         & 34.1   & 63.2    & 36.5   & 63.2      & \textbf{39.2}    & \textbf{64.1}     \\ 
kitchen            & 49.7   & 78.7    & 51.0   & 79.8      & \textbf{59.1}    & \textbf{83.1}     \\ 
meeting room       & 49.1   & 78.1    & 57.8   & 90.5      & \textbf{58.2}    & \textbf{90.8}     \\ 
classroom-2        & 47.7   & 79.5    & 50.1   & 80.4      & \textbf{50.9}    & \textbf{81.9}     \\ \midrule
Average            & 44.9   & 74.1    & 45.8   & 76.7      & \textbf{51.3}    & \textbf{79.7}    \\
\bottomrule
\end{tabular}
}
\setlength{\tabcolsep}{6pt}
\vspace{-1mm}
\caption{Effectiveness of using region-specific semantic features, context-aware semantic features, and both.}
\label{tab:Dual-semantic Features}
\vspace{-2mm}
\end{table}

\begin{figure}[t]
\centering
\includegraphics[width=\linewidth]{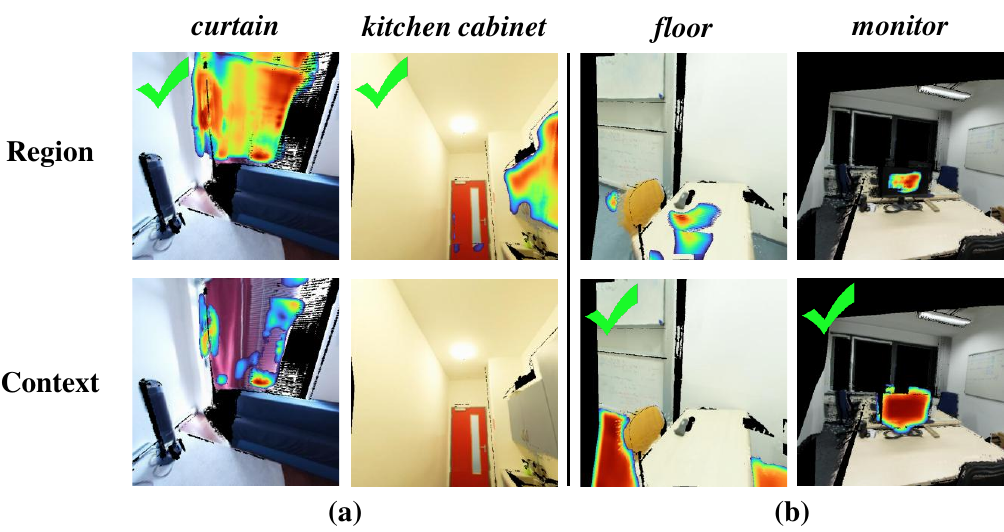} \\
\vspace{-1mm}
\caption{Examples on (a) cases where region-specific features are selected and (b) cases where context-specific features are selected.}
\vspace{-4mm}
\label{fig:Head Selection}
\end{figure}

We analyze rendered views to observe when region-specific versus context-aware features are selected in \ours. As examples in Fig.~\ref{fig:Head Selection}, the trends align with intuition: for objects or surfaces hard to identify in isolation (like monitors or floors), context-aware features typically show stronger responses and are selected. For distinguishable objects observed in isolation (such as cabinets), region-specific features often have higher responses, likely due to minimal interference from neighboring regions.

\noindent\textbf{Ablation on Feature Selection Strategies} We compare the performance of using another strategy and the upper bound. Our used strategy achieves higher performance, approximating the upper bound performance. Please refer to Supplementary for more details.

\noindent\textbf{Ablation on Feature Dimension $K$} Please see the analysis in our Supplementary.

\noindent\textbf{Ablation on Threshold of Relevance Score} We study its influence in our Supplementary. 

\subsection{Discussion}
\label{sec:discussion}
\begin{figure}[t]
\vspace{-2mm}
\centering
\includegraphics[width=\linewidth]{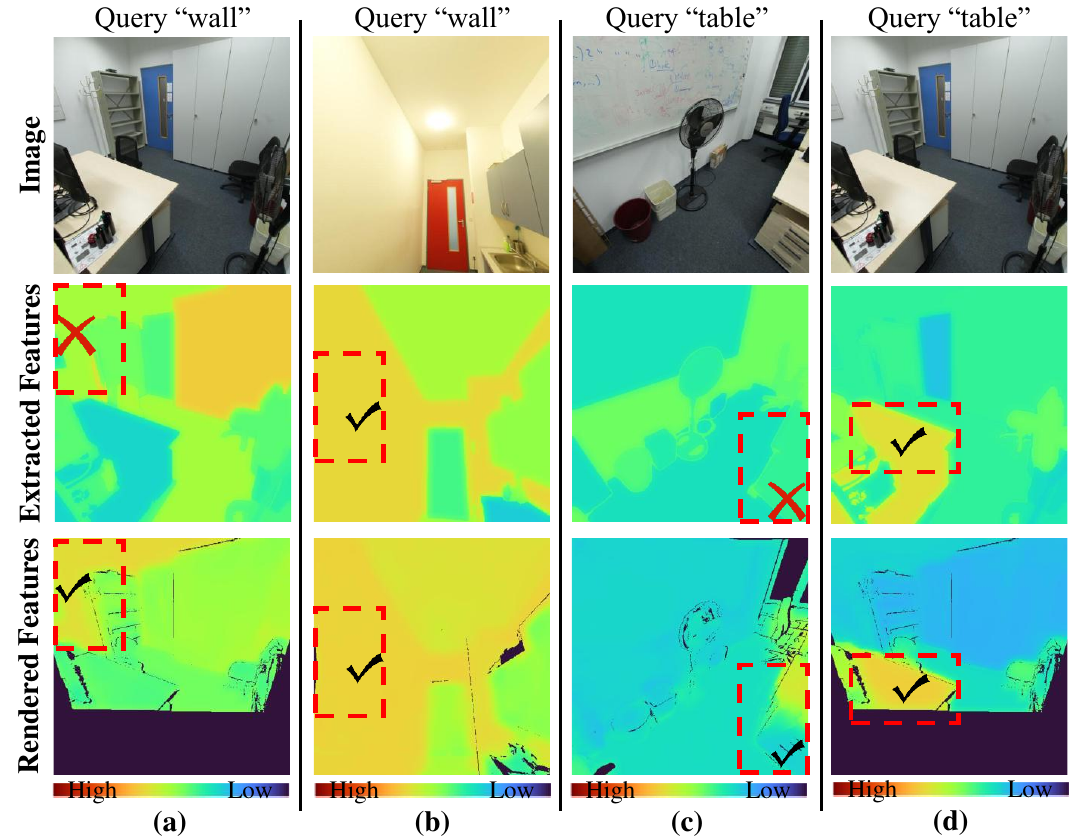} \\
\vspace{-1mm}
\caption{(a) and (c) show two examples that the extracted 2D supervision features are inaccurate for ``wall" and ``table" regions (see the second row), and (b) and (d) show two examples that the supervision features are accurate. Our model is robust to noisy labels and can learn accurate features (see the third row). }
\vspace{-3mm}
\label{fig:discussion}
\end{figure}

Our \ourss outperforms the per-scene-based optimization method LangSplat in semantic field modeling, exceeding our initial expectations. As we know, due to dense input views and per-scene-based optimization, the novel view synthesize quality (\ieno, RGB quality) of LangSplat is much better than that of Splatt3R and our \ours. Why the semantic modeling capability of our \ourss is better than LangSplat? We attribute the reason for the robustness of our framework to the interference of noisy semantic features. As some examples in Fig.~\ref{fig:discussion}, the extracted 2D semantic features (that are used as supervision) are inaccurate and some regions (such as the ``wall" in (a) and ``table" in (c) cannot be correctly identified when querying with texts. LangSplat tends to overfit these unreliable features. In contrast, our model learns the semantics across different scenes where consistent and reliable patterns are learned, being robust to the noise. The learned features (as shown in the third row in Fig.~\ref{fig:discussion}) respond correctly to those queries even though the supervision features are wrong.      

\section{Conclusion}
In this paper, we introduced a new framework, \ours, for generalizable 3D semantic field modeling using sparse, uncalibrated images. Unlike existing methods that rely on dense, calibrated images and costly per-scene optimization, \ourss allows rapid inference of semantic 3D Gaussians of a new scene in a single feed-forward pass from just two uncalibrated images. \ourss demonstrates superior semantic understanding performance compared to the representative per-scene optimization-based method LangSplat, achieving 4000 $\times$ faster. 
Our dual-feature approach enhances the reliability of semantic features without human annotations. But, there is still a gap between the ``ground truth" semantics.  
We hope this work encourages more investigations to advance the generalizable 3D understanding. 

{
    \small
    \bibliographystyle{ieeenat_fullname}
    \bibliography{main}

\begin{thebibliography}{50}
\providecommand{\natexlab}[1]{#1}
\providecommand{\url}[1]{\texttt{#1}}
\expandafter\ifx\csname urlstyle\endcsname\relax
  \providecommand{\doi}[1]{doi: #1}\else
  \providecommand{\doi}{doi: \begingroup \urlstyle{rm}\Url}\fi

\bibitem[Bhalgat et~al.(2024)Bhalgat, Laina, Henriques, Zisserman, and Vedaldi]{bhalgat2024n2f2}
Yash Bhalgat, Iro Laina, Jo{\~a}o~F Henriques, Andrew Zisserman, and Andrea Vedaldi.
\newblock {N2F2}: Hierarchical scene understanding with nested neural feature fields.
\newblock \emph{arXiv preprint arXiv:2403.10997}, 2024.

\bibitem[Charatan et~al.(2024)Charatan, Li, Tagliasacchi, and Sitzmann]{charatan2024pixelsplat}
David Charatan, Sizhe~Lester Li, Andrea Tagliasacchi, and Vincent Sitzmann.
\newblock pixelsplat: 3d gaussian splats from image pairs for scalable generalizable 3d reconstruction.
\newblock In \emph{CVPR}, pages 19457--19467, 2024.

\bibitem[Chen et~al.(2021)Chen, Xu, Zhao, Zhang, Xiang, Yu, and Su]{chen2021mvsnerf}
Anpei Chen, Zexiang Xu, Fuqiang Zhao, Xiaoshuai Zhang, Fanbo Xiang, Jingyi Yu, and Hao Su.
\newblock Mvsnerf: Fast generalizable radiance field reconstruction from multi-view stereo.
\newblock In \emph{ICCV}, pages 14124--14133, 2021.

\bibitem[Chen et~al.(2024)Chen, Xu, Zheng, Zhuang, Pollefeys, Geiger, Cham, and Cai]{chen2024mvsplat}
Yuedong Chen, Haofei Xu, Chuanxia Zheng, Bohan Zhuang, Marc Pollefeys, Andreas Geiger, Tat-Jen Cham, and Jianfei Cai.
\newblock Mvsplat: Efficient 3d gaussian splatting from sparse multi-view images.
\newblock In \emph{ECCV}, 2024.

\bibitem[Cheng et~al.(2024)Cheng, Long, Yang, Yao, Yin, Ma, Wang, and Chen]{cheng2024gaussianpro}
Kai Cheng, Xiaoxiao Long, Kaizhi Yang, Yao Yao, Wei Yin, Yuexin Ma, Wenping Wang, and Xuejin Chen.
\newblock Gaussianpro: 3d gaussian splatting with progressive propagation.
\newblock In \emph{ICML}, 2024.

\bibitem[Dai et~al.(2017)Dai, Chang, Savva, Halber, Funkhouser, and Nie{\ss}ner]{dai2017scannet}
Angela Dai, Angel~X Chang, Manolis Savva, Maciej Halber, Thomas Funkhouser, and Matthias Nie{\ss}ner.
\newblock Scannet: Richly-annotated 3d reconstructions of indoor scenes.
\newblock In \emph{CVPR}, pages 5828--5839, 2017.

\bibitem[Dong et~al.(2023)Dong, Bao, Zheng, Zhang, Chen, Yang, Zeng, Zhang, Yuan, Chen, et~al.]{dong2023maskclip}
Xiaoyi Dong, Jianmin Bao, Yinglin Zheng, Ting Zhang, Dongdong Chen, Hao Yang, Ming Zeng, Weiming Zhang, Lu Yuan, Dong Chen, et~al.
\newblock Maskclip: Masked self-distillation advances contrastive language-image pretraining.
\newblock In \emph{CVPR}, pages 10995--11005, 2023.

\bibitem[Fan et~al.(2024)Fan, Cong, Wen, Wang, Zhang, Ding, Xu, Ivanovic, Pavone, Pavlakos, et~al.]{fan2024instantsplat}
Zhiwen Fan, Wenyan Cong, Kairun Wen, Kevin Wang, Jian Zhang, Xinghao Ding, Danfei Xu, Boris Ivanovic, Marco Pavone, Georgios Pavlakos, et~al.
\newblock Instantsplat: Unbounded sparse-view pose-free gaussian splatting in 40 seconds.
\newblock \emph{arXiv preprint arXiv:2403.20309}, 2, 2024.

\bibitem[Fei et~al.(2024)Fei, Zheng, Duan, Zhan, Tomizuka, Keutzer, and Lu]{fei2024pixelgaussian}
Xin Fei, Wenzhao Zheng, Yueqi Duan, Wei Zhan, Masayoshi Tomizuka, Kurt Keutzer, and Jiwen Lu.
\newblock Pixelgaussian: Generalizable 3d gaussian reconstruction from arbitrary views.
\newblock \emph{arXiv preprint arXiv:2410.18979}, 2024.

\bibitem[Fu et~al.(2024)Fu, Liu, Kulkarni, Kautz, Efros, and Wang]{Fu_2024_CVPR}
Yang Fu, Sifei Liu, Amey Kulkarni, Jan Kautz, Alexei~A. Efros, and Xiaolong Wang.
\newblock Colmap-free 3d gaussian splatting.
\newblock In \emph{CVPR}, pages 20796--20805, 2024.

\bibitem[Hu et~al.(2024)Hu, Wang, Fan, Fan, Peng, Lei, Li, and Zhang]{hu2024semantic}
Xu Hu, Yuxi Wang, Lue Fan, Junsong Fan, Junran Peng, Zhen Lei, Qing Li, and Zhaoxiang Zhang.
\newblock Semantic anything in 3d gaussians.
\newblock \emph{arXiv preprint arXiv:2401.17857}, 2024.

\bibitem[Jain et~al.(2021)Jain, Tancik, and Abbeel]{jain2021putting}
Ajay Jain, Matthew Tancik, and Pieter Abbeel.
\newblock Putting nerf on a diet: Semantically consistent few-shot view synthesis.
\newblock In \emph{ICCV}, pages 5885--5894, 2021.

\bibitem[Ji et~al.(2024)Ji, Zhu, Tang, Liu, Zhang, Xie, Ma, and Tan]{ji2024fastlgs}
Yuzhou Ji, He Zhu, Junshu Tang, Wuyi Liu, Zhizhong Zhang, Yuan Xie, Lizhuang Ma, and Xin Tan.
\newblock Fastlgs: Speeding up language embedded gaussians with feature grid mapping.
\newblock \emph{arXiv preprint arXiv:2406.01916}, 2024.

\bibitem[Kerbl et~al.(2023)Kerbl, Kopanas, Leimk{\"u}hler, and Drettakis]{kerbl20233d}
Bernhard Kerbl, Georgios Kopanas, Thomas Leimk{\"u}hler, and George Drettakis.
\newblock 3d gaussian splatting for real-time radiance field rendering.
\newblock \emph{TOG}, 42\penalty0 (4):\penalty0 139--1, 2023.

\bibitem[Kerr et~al.(2023)Kerr, Kim, Goldberg, Kanazawa, and Tancik]{kerr2023lerf}
Justin Kerr, Chung~Min Kim, Ken Goldberg, Angjoo Kanazawa, and Matthew Tancik.
\newblock Lerf: Language embedded radiance fields.
\newblock In \emph{ICCV}, pages 19729--19739, 2023.

\bibitem[Kirillov et~al.(2023)Kirillov, Mintun, Ravi, Mao, Rolland, Gustafson, Xiao, Whitehead, Berg, Lo, et~al.]{kirillov2023segment}
Alexander Kirillov, Eric Mintun, Nikhila Ravi, Hanzi Mao, Chloe Rolland, Laura Gustafson, Tete Xiao, Spencer Whitehead, Alexander~C Berg, Wan-Yen Lo, et~al.
\newblock Segment anything.
\newblock In \emph{CVPR}, pages 4015--4026, 2023.

\bibitem[Leroy et~al.(2024)Leroy, Cabon, and Revaud]{leroy2024grounding}
Vincent Leroy, Yohann Cabon, and J{\'e}r{\^o}me Revaud.
\newblock Grounding image matching in 3d with mast3r.
\newblock \emph{arXiv preprint arXiv:2406.09756}, 2024.

\bibitem[Liao et~al.(2024)Liao, Li, Bao, Ye, Wang, Li, and Liu]{liao2024clip}
Guibiao Liao, Jiankun Li, Zhenyu Bao, Xiaoqing Ye, Jingdong Wang, Qing Li, and Kanglin Liu.
\newblock Clip-gs: Clip-informed gaussian splatting for real-time and view-consistent 3d semantic understanding.
\newblock \emph{arXiv preprint arXiv:2404.14249}, 2024.

\bibitem[Liu et~al.(2023)Liu, Zhan, Zhang, Xu, Yu, El~Saddik, Theobalt, Xing, and Lu]{liu2023weakly}
Kunhao Liu, Fangneng Zhan, Jiahui Zhang, Muyu Xu, Yingchen Yu, Abdulmotaleb El~Saddik, Christian Theobalt, Eric Xing, and Shijian Lu.
\newblock Weakly supervised 3d open-vocabulary segmentation.
\newblock \emph{NeurIPS}, 36:\penalty0 53433--53456, 2023.

\bibitem[Lu et~al.(2024)Lu, Yu, Xu, Xiangli, Wang, Lin, and Dai]{lu2024scaffold}
Tao Lu, Mulin Yu, Linning Xu, Yuanbo Xiangli, Limin Wang, Dahua Lin, and Bo Dai.
\newblock Scaffold-gs: Structured 3d gaussians for view-adaptive rendering.
\newblock In \emph{CVPR}, pages 20654--20664, 2024.

\bibitem[Mildenhall et~al.(2021)Mildenhall, Srinivasan, Tancik, Barron, Ramamoorthi, and Ng]{mildenhall2021nerf}
Ben Mildenhall, Pratul~P Srinivasan, Matthew Tancik, Jonathan~T Barron, Ravi Ramamoorthi, and Ren Ng.
\newblock Nerf: Representing scenes as neural radiance fields for view synthesis.
\newblock \emph{CACM}, 65\penalty0 (1):\penalty0 99--106, 2021.

\bibitem[Niemeyer et~al.(2022)Niemeyer, Barron, Mildenhall, Sajjadi, Geiger, and Radwan]{niemeyer2022regnerf}
Michael Niemeyer, Jonathan~T Barron, Ben Mildenhall, Mehdi~SM Sajjadi, Andreas Geiger, and Noha Radwan.
\newblock Regnerf: Regularizing neural radiance fields for view synthesis from sparse inputs.
\newblock In \emph{CVPR}, pages 5480--5490, 2022.

\bibitem[Niemeyer et~al.(2024)Niemeyer, Manhardt, Rakotosaona, Oechsle, Duckworth, Gosula, Tateno, Bates, Kaeser, and Tombari]{niemeyer2024radsplat}
Michael Niemeyer, Fabian Manhardt, Marie-Julie Rakotosaona, Michael Oechsle, Daniel Duckworth, Rama Gosula, Keisuke Tateno, John Bates, Dominik Kaeser, and Federico Tombari.
\newblock Radsplat: Radiance field-informed gaussian splatting for robust real-time rendering with 900+ fps.
\newblock \emph{arXiv preprint arXiv:2403.13806}, 2024.

\bibitem[Oquab et~al.(2023)Oquab, Darcet, Moutakanni, Vo, Szafraniec, Khalidov, Fernandez, Haziza, Massa, El-Nouby, et~al.]{oquab2023dinov2}
Maxime Oquab, Timoth{\'e}e Darcet, Th{\'e}o Moutakanni, Huy Vo, Marc Szafraniec, Vasil Khalidov, Pierre Fernandez, Daniel Haziza, Francisco Massa, Alaaeldin El-Nouby, et~al.
\newblock Dinov2: Learning robust visual features without supervision.
\newblock \emph{arXiv preprint arXiv:2304.07193}, 2023.

\bibitem[Qin et~al.(2024)Qin, Li, Zhou, Wang, and Pfister]{qin2024langsplat}
Minghan Qin, Wanhua Li, Jiawei Zhou, Haoqian Wang, and Hanspeter Pfister.
\newblock Langsplat: 3d language gaussian splatting.
\newblock In \emph{CVPR}, pages 20051--20060, 2024.

\bibitem[Qiu et~al.(2024)Qiu, Yang, Zeng, and Wang]{qiu2024feature}
Ri-Zhao Qiu, Ge Yang, Weijia Zeng, and Xiaolong Wang.
\newblock Feature splatting: Language-driven physics-based scene synthesis and editing.
\newblock \emph{arXiv preprint arXiv:2404.01223}, 2024.

\bibitem[Qu et~al.(2024)Qu, Dai, Li, Lin, Cao, Zhang, and Ji]{qu2024goi}
Yansong Qu, Shaohui Dai, Xinyang Li, Jianghang Lin, Liujuan Cao, Shengchuan Zhang, and Rongrong Ji.
\newblock Goi: Find 3d gaussians of interest with an optimizable open-vocabulary semantic-space hyperplane.
\newblock \emph{arXiv preprint arXiv:2405.17596}, 2024.

\bibitem[Radford et~al.(2021)Radford, Kim, Hallacy, Ramesh, Goh, Agarwal, Sastry, Askell, Mishkin, Clark, et~al.]{radford2021learning}
Alec Radford, Jong~Wook Kim, Chris Hallacy, Aditya Ramesh, Gabriel Goh, Sandhini Agarwal, Girish Sastry, Amanda Askell, Pamela Mishkin, Jack Clark, et~al.
\newblock Learning transferable visual models from natural language supervision.
\newblock In \emph{ICML}, pages 8748--8763. PMLR, 2021.

\bibitem[Schmidt et~al.(2024)Schmidt, Piekenbrinck, and Leibe]{schmidt2024look}
Christian Schmidt, Jens Piekenbrinck, and Bastian Leibe.
\newblock Look gauss, no pose: Novel view synthesis using gaussian splatting without accurate pose initialization.
\newblock \emph{arXiv preprint arXiv:2410.08743}, 2024.

\bibitem[Seo et~al.(2023{\natexlab{a}})Seo, Chang, and Kwak]{seo2023flipnerf}
Seunghyeon Seo, Yeonjin Chang, and Nojun Kwak.
\newblock Flipnerf: Flipped reflection rays for few-shot novel view synthesis.
\newblock In \emph{ICCV}, pages 22883--22893, 2023{\natexlab{a}}.

\bibitem[Seo et~al.(2023{\natexlab{b}})Seo, Han, Chang, and Kwak]{seo2023mixnerf}
Seunghyeon Seo, Donghoon Han, Yeonjin Chang, and Nojun Kwak.
\newblock Mixnerf: Modeling a ray with mixture density for novel view synthesis from sparse inputs.
\newblock In \emph{CVPR}, pages 20659--20668, 2023{\natexlab{b}}.

\bibitem[Shafiullah et~al.(2022)Shafiullah, Paxton, Pinto, Chintala, and Szlam]{shafiullah2022clip}
Nur Muhammad~Mahi Shafiullah, Chris Paxton, Lerrel Pinto, Soumith Chintala, and Arthur Szlam.
\newblock Clip-fields: Weakly supervised semantic fields for robotic memory.
\newblock \emph{arXiv preprint arXiv:2210.05663}, 2022.

\bibitem[Shen et~al.(2023)Shen, Yang, Yu, Wong, Kaelbling, and Isola]{shen2023distilled}
William Shen, Ge Yang, Alan Yu, Jansen Wong, Leslie~Pack Kaelbling, and Phillip Isola.
\newblock Distilled feature fields enable few-shot language-guided manipulation.
\newblock \emph{arXiv preprint arXiv:2308.07931}, 2023.

\bibitem[Shi et~al.(2024)Shi, Wang, Duan, and Guan]{shi2024language}
Jin-Chuan Shi, Miao Wang, Hao-Bin Duan, and Shao-Hua Guan.
\newblock Language embedded 3d gaussians for open-vocabulary scene understanding.
\newblock In \emph{CVPR}, pages 5333--5343, 2024.

\bibitem[Smart et~al.(2024)Smart, Zheng, Laina, and Prisacariu]{smart2024splatt3r}
Brandon Smart, Chuanxia Zheng, Iro Laina, and Victor~Adrian Prisacariu.
\newblock Splatt3r: Zero-shot gaussian splatting from uncalibarated image pairs.
\newblock \emph{arXiv preprint arXiv:2408.13912}, 2024.

\bibitem[Szymanowicz et~al.(2024{\natexlab{a}})Szymanowicz, Insafutdinov, Zheng, Campbell, Henriques, Rupprecht, and Vedaldi]{szymanowicz2024flash3d}
Stanislaw Szymanowicz, Eldar Insafutdinov, Chuanxia Zheng, Dylan Campbell, Jo{\~a}o~F Henriques, Christian Rupprecht, and Andrea Vedaldi.
\newblock Flash3d: Feed-forward generalisable 3d scene reconstruction from a single image.
\newblock \emph{arXiv preprint arXiv:2406.04343}, 2024{\natexlab{a}}.

\bibitem[Szymanowicz et~al.(2024{\natexlab{b}})Szymanowicz, Rupprecht, and Vedaldi]{szymanowicz2024splatter}
Stanislaw Szymanowicz, Chrisitian Rupprecht, and Andrea Vedaldi.
\newblock Splatter image: Ultra-fast single-view 3d reconstruction.
\newblock In \emph{CVPR}, pages 10208--10217, 2024{\natexlab{b}}.

\bibitem[Tang et~al.(2025)Tang, Chen, Chen, Wang, Zeng, and Liu]{tang2025lgm}
Jiaxiang Tang, Zhaoxi Chen, Xiaokang Chen, Tengfei Wang, Gang Zeng, and Ziwei Liu.
\newblock Lgm: Large multi-view gaussian model for high-resolution 3d content creation.
\newblock In \emph{ECCV}, pages 1--18. Springer, 2025.

\bibitem[Wang et~al.(2021)Wang, Wang, Genova, Srinivasan, Zhou, Barron, Martin-Brualla, Snavely, and Funkhouser]{wang2021ibrnet}
Qianqian Wang, Zhicheng Wang, Kyle Genova, Pratul~P Srinivasan, Howard Zhou, Jonathan~T Barron, Ricardo Martin-Brualla, Noah Snavely, and Thomas Funkhouser.
\newblock Ibrnet: Learning multi-view image-based rendering.
\newblock In \emph{CVPR}, pages 4690--4699, 2021.

\bibitem[Wang et~al.(2024{\natexlab{a}})Wang, Leroy, Cabon, Chidlovskii, and Revaud]{wang2024dust3r}
Shuzhe Wang, Vincent Leroy, Yohann Cabon, Boris Chidlovskii, and Jerome Revaud.
\newblock Dust3r: Geometric 3d vision made easy.
\newblock In \emph{CVPR}, pages 20697--20709, 2024{\natexlab{a}}.

\bibitem[Wang et~al.(2024{\natexlab{b}})Wang, Huang, Chen, and Lee]{wang2024freesplat}
Yunsong Wang, Tianxin Huang, Hanlin Chen, and Gim~Hee Lee.
\newblock Freesplat: Generalizable 3d gaussian splatting towards free-view synthesis of indoor scenes.
\newblock \emph{arXiv preprint arXiv:2405.17958}, 2024{\natexlab{b}}.

\bibitem[Wewer et~al.(2024)Wewer, Raj, Ilg, Schiele, and Lenssen]{wewer2024latentsplat}
Christopher Wewer, Kevin Raj, Eddy Ilg, Bernt Schiele, and Jan~Eric Lenssen.
\newblock latentsplat: Autoencoding variational gaussians for fast generalizable 3d reconstruction.
\newblock \emph{arXiv preprint arXiv:2403.16292}, 2024.

\bibitem[Xu et~al.(2024{\natexlab{a}})Xu, Chen, Chen, Zhang, Yu, and Yang]{xu2024tiger}
Teng Xu, Jiamin Chen, Peng Chen, Youjia Zhang, Junqing Yu, and Wei Yang.
\newblock Tiger: Text-instructed 3d gaussian retrieval and coherent editing.
\newblock \emph{arXiv preprint arXiv:2405.14455}, 2024{\natexlab{a}}.

\bibitem[Xu et~al.(2024{\natexlab{b}})Xu, Shi, Yifan, Chen, Yang, Peng, Shen, and Wetzstein]{xu2024grm}
Yinghao Xu, Zifan Shi, Wang Yifan, Hansheng Chen, Ceyuan Yang, Sida Peng, Yujun Shen, and Gordon Wetzstein.
\newblock Grm: Large gaussian reconstruction model for efficient 3d reconstruction and generation.
\newblock \emph{arXiv preprint arXiv:2403.14621}, 2024{\natexlab{b}}.

\bibitem[Yeshwanth et~al.(2023)Yeshwanth, Liu, Nie{\ss}ner, and Dai]{yeshwanth2023scannet++}
Chandan Yeshwanth, Yueh-Cheng Liu, Matthias Nie{\ss}ner, and Angela Dai.
\newblock Scannet++: A high-fidelity dataset of 3d indoor scenes.
\newblock In \emph{ICCV}, pages 12--22, 2023.

\bibitem[Yu et~al.(2024)Yu, Chen, Huang, Sattler, and Geiger]{yu2024mip}
Zehao Yu, Anpei Chen, Binbin Huang, Torsten Sattler, and Andreas Geiger.
\newblock Mip-splatting: Alias-free 3d gaussian splatting.
\newblock In \emph{CVPR}, pages 19447--19456, 2024.

\bibitem[Zhang et~al.(2024)Zhang, Zou, Li, Yi, and Wang]{zhang2024transplat}
Chuanrui Zhang, Yingshuang Zou, Zhuoling Li, Minmin Yi, and Haoqian Wang.
\newblock Transplat: Generalizable 3d gaussian splatting from sparse multi-view images with transformers.
\newblock \emph{arXiv preprint arXiv:2408.13770}, 2024.

\bibitem[Zhang et~al.(2025)Zhang, Bi, Tan, Xiangli, Zhao, Sunkavalli, and Xu]{zhang2025gs}
Kai Zhang, Sai Bi, Hao Tan, Yuanbo Xiangli, Nanxuan Zhao, Kalyan Sunkavalli, and Zexiang Xu.
\newblock Gs-lrm: Large reconstruction model for 3d gaussian splatting.
\newblock In \emph{ECCV}, pages 1--19. Springer, 2025.

\bibitem[Zhou et~al.(2024)Zhou, Chang, Jiang, Fan, Zhu, Xu, Chari, You, Wang, and Kadambi]{zhou2024feature}
Shijie Zhou, Haoran Chang, Sicheng Jiang, Zhiwen Fan, Zehao Zhu, Dejia Xu, Pradyumna Chari, Suya You, Zhangyang Wang, and Achuta Kadambi.
\newblock Feature {3DGS}: Supercharging 3d gaussian splatting to enable distilled feature fields.
\newblock In \emph{CVPR}, pages 21676--21685, 2024.

\bibitem[Zhu et~al.(2024)Zhu, He, Li, Li, and Chen]{zhu2024vanilla}
Hanxin Zhu, Tianyu He, Xin Li, Bingchen Li, and Zhibo Chen.
\newblock Is vanilla mlp in neural radiance field enough for few-shot view synthesis?
\newblock In \emph{CVPR}, pages 20288--20298, 2024.

\end{thebibliography}
}

\clearpage
\setcounter{page}{1}
\maketitlesupplementary

\section{Effectiveness of Feature Selection Strategy}

It is difficult to determine which feature to choose (from the region-specific and context-aware features) for each querying. 
We follow the strategy used in LERF \cite{kerr2023lerf} to choose the feature that yields a higher regularized relevancy score for each query. Our statistical analysis in Tab.~\ref{tab:accuracy of strategy} shows that this strategy can hit better features at an average accuracy of 70\% when we do the study on the extracted 2D semantic features (which are used as supervision of the rendered features), being much better than a random guess. A higher relevance score of the features usually indicates a higher correlation with the querying text and thus the features are prone to be the right one. 

Particularly, for each scene, 100 samples are randomly selected for the statistical analysis. For each query, we compute the IoU of region-specific and context-aware features, respectively. The one with a higher IoU is considered as the ``ground truth" correct selection of feature type for the current query over the sample. Finally, we assess how often this strategy hits the better-performing feature, and Tab.~\ref{tab:accuracy of strategy} shows that the average accuracy across all the scenes reaches 70\%, indicating the effectiveness of this strategy. Note that we focus only on cases where the IoU difference between the two types of features exceeds 5\% to exclude the cases where the two types of features have similar performance.

\begin{table}[h]
    \centering
    \small
    \setlength{\tabcolsep}{9mm}
    \begin{tabular}{c|c}
    \toprule
         Scene  & Accuracy\\ \midrule
         office  & 68.5\\
         classroom-1  & 82.2 \\
         utility room  & 71.8 \\
         workshop  & 75.6\\
         printing room  & 65.2\\
         kitchen  & 77.0 \\
         meeting room  & 62.2\\
         classroom-2  & 56.3 \\ \midrule
         Average  & 69.9 \\
    \bottomrule
    \end{tabular}
     \caption{Effectiveness of our used feature selection strategy evaluated on the extracted 2D semantic features.}
    \label{tab:accuracy of strategy}
\end{table}

\begin{figure*}[!h]
\centering
\includegraphics[width=\linewidth]{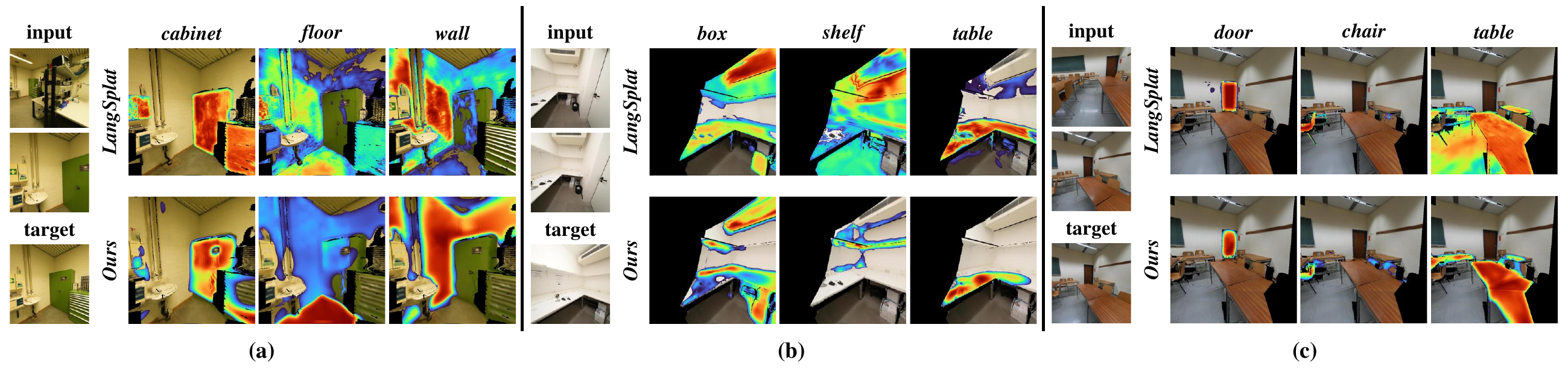} \\
\caption{Additional qualitative comparison between LangSplat and our \ours.}
\vspace{-1mm}
\label{fig:additional quantitative comparison}
\end{figure*}
\section{More Visualization Comparison}
We provide more visualization comparison of results between LangSplat and \ours. As illustrated in Fig.~\ref{fig:additional quantitative comparison} (a), LangSplat faces obvious confusion between ``wall'' and ``floor'', while \ourss successfully distinguishes these categories. In Fig.~\ref{fig:additional quantitative comparison}~(b), when querying ``box'', both methods show suboptimal performance, where \ours produces a stronger response for the box in the bottom-right corner. Moreover, for the narrow ``shelf'', our method achieves a more precise localization while LangSplat produces high responses on the ``wall" region. In Fig.~\ref{fig:additional quantitative comparison}~(c), LangSplat mistakenly trades certain ``floor" region as ``table" while ours produces more accurate localization. 

\section{More Visualization on Generalization Results} We provide the visualization of testing on scenes from other datasets in the main manuscript. Due to space limitations, we only show the rendered results on one target view. In our supplementary, we provide the results on more rendering views in the form of videos. Please see the attached zip folder.

\section{Ablation Studies}

\noindent\textbf{Effectiveness of Dual Features} In Tab.~\ref{tab:Dual-semantic Features} of our main manuscript, we statically investigated the effectiveness of using dual feature design by randomly sampling 100 samples for each scene for testing. 

Here, we also evaluate the results when using the same setting as that in Tab.~\ref{tab:Comparison Results} where only a sample for each scene is tested. Tab.~\ref{tab:Dual-semantic Features One Sample} shows the results. We obtain the consistent conclusion that combining region-specific and context-aware features leads to superior performance in most of the scenes. Note that the statistical results in Tab.~\ref{tab:Dual-semantic Features} are more reliable considering that they can eliminate the influence of randomness caused by the small amount of the testing data.      

\begin{table}[!h]
\centering
\small
\setlength{\tabcolsep}{1mm}
\resizebox{0.42\textwidth}{!}{ 
\begin{tabular}{c|cc|cc|cc}
\toprule
\multirow{2}{*}{Scene} & \multicolumn{2}{c|}{Region} & \multicolumn{2}{c|}{Context} & \multicolumn{2}{c}{Both} \\ \cmidrule{2-7}
                       & mIoU$\uparrow$     & LA$\uparrow$        & mIoU$\uparrow$     & LA$\uparrow$          & mIoU$\uparrow$      & LA$\uparrow$   \\  \midrule
office             & 78.6   & 83.3    & 65.8   & 83.3      & \textbf{86.1}    & \textbf{100}   \\  
classroom-1        & 71.1   & 66.7    & \textbf{79.1}   & 100       & 78.9    & \textbf{100}  \\ 
utility room       & 78.9   & 100     & 46.8   & 100      & \textbf{78.9}    & \textbf{100}     \\ 
workshop           & 59.4   & 100     & 70.2   & 100      & \textbf{76.0}    & \textbf{100}     \\ 
printing room      & 38.8   & 80.0    & 51.1   & 60.0      & \textbf{53.4}   & \textbf{80.0}     \\ 
kitchen            & 82.9   & 100    & \textbf{88.8}   & 100      & 84.1   & \textbf{100}     \\ 
meeting room       & 57.8   & 83.3    & 68.4   & 100      & \textbf{73.6}   & \textbf{100}     \\ 
classroom-2        & 67.4   & 100    & 72.0   & 100      & \textbf{73.8}   & \textbf{100}     \\ \midrule
Average            & 66.9   & 89.2    & 67.8   & 92.9      & \textbf{75.6}   & \textbf{97.5}    \\
\bottomrule
\end{tabular}
}
\setlength{\tabcolsep}{6pt}
\vspace{-1mm}
\caption{Effectiveness of using region-specific semantic features, context-aware semantic features, and both when the setting is same with Tab.~\ref{tab:Comparison Results} where only a sample for each scene is tested.}
\label{tab:Dual-semantic Features One Sample}
\vspace{-2mm}
\end{table}

Moreover, we test on other settings with at least 30\%, 70\%, and 90\% pixel correspondence between input-input and target-input image pairs, whereas the main manuscript only shows the results for 50\% pixel correspondence (in Sec. \ref{sec:ablation studies}). Similarly, we randomly sample 100 samples of each scene. The final results are shown in Tab.~\ref{tab:other correspondence}, which demonstrates the effectiveness of using both region-specific and context-aware features.

\begin{table*}[t]
\centering
\normalsize
\small
\setlength{\tabcolsep}{0.45mm}
\begin{tabular}{c|cc|cc|cc|cc|cc|cc|cc|cc|cc}
\toprule
\multirow{3}{*}{Scene} & \multicolumn{6}{c|}{0.3}                                                             & \multicolumn{6}{c|}{0.7}                                                             & \multicolumn{6}{c}{0.9}                                                             \\ \cmidrule{2-19}
                       & \multicolumn{2}{c|}{Region} & \multicolumn{2}{c|}{Context} & \multicolumn{2}{c|}{Both} & \multicolumn{2}{c|}{Region} & \multicolumn{2}{c|}{Context} & \multicolumn{2}{c|}{Both} & \multicolumn{2}{c|}{Region} & \multicolumn{2}{c|}{Context} & \multicolumn{2}{c}{Both} \\ \cmidrule{2-19}
                       & mIoU$\uparrow$         & LA$\uparrow$          & mIoU$\uparrow$         & LA$\uparrow$           & mIoU$\uparrow$        & LA$\uparrow$         & mIoU$\uparrow$         & LA$\uparrow$          & mIoU$\uparrow$         & LA$\uparrow$           & mIoU$\uparrow$        & LA$\uparrow$         & mIoU$\uparrow$         & LA$\uparrow$          & mIoU$\uparrow$         & LA$\uparrow$           & mIoU$\uparrow$        & LA$\uparrow$         \\ \midrule
office             & 46.0       & 72.1      & 46.4       & 76.4       & \textbf{51.4}      & \textbf{79.5}     & 52.5       & 79.0      & 52.7       & 81.1       & \textbf{58.3}      & \textbf{87.4}     & 54.4       & 78.3      & 50.1       & 81.6       & \textbf{58.6}      & \textbf{85.6}     \\
classroom-1             & 43.1       & 68.5      & 44.6       & 76.0       & \textbf{47.7}      & \textbf{76.7}    & 45.3       & 67.8      & 46.6       & \textbf{76.6}       & \textbf{49.7}      & 76.3     & 50.9       & 76.0      & 54.9       & 88.0       & \textbf{57.1}      & \textbf{88.2}     \\
utility room             & 42.3       & 68.1      & 36.2       & 66.3       & \textbf{48.5}      & \textbf{76.9}     & 50.7       & 80.2      & 42.7       & 79.0       & \textbf{55.3}      & \textbf{85.9}     & 60.0       & 90.3      & 48.8       & \textbf{92.7}       & \textbf{63.2}      & 92.4     \\
workshop             & 36.4       & 73.8      & 33.7       & 65.0       & \textbf{43.3}      & \textbf{75.6}     & 40.2       & 74.8      & 36.5       & 64.9       & \textbf{48.1}      & \textbf{75.9}     & 49.5       & 79.5      & 44.3       & 74.7       & \textbf{55.9}      & \textbf{82.2}     \\
printing room             & 29.7       & 58.2      & 34.1       & \textbf{61.7}       & \textbf{35.5}      & 60.7     & 34.0       & 63.2      & 35.9       & 63.5       & \textbf{38.2}      & \textbf{65.3}     & 32.9       & 63.2      & 35.9       & 61.7       & \textbf{37.5}      & \textbf{64.8}     \\
kitchen             & 44.0       & 75.2      & 43.0       & 71.2       & \textbf{50.6}      & \textbf{78.4}     & 52.9       & 85.0      & 55.0       & 82.8       & \textbf{63.7}      & \textbf{86.2}     & 58.1       & 85.5      & 57.1       & 84.3       & \textbf{64.5}      & \textbf{87.3}     \\
meeting room             & 45.3       & 74.6      & 53.3       & \textbf{89.3}       & \textbf{53.6}      & 87.9     & 50.3       & 81.2      & \textbf{61.9}       & 92.5       & 61.2      & \textbf{92.6}     & 49.5       & 81.8      & \textbf{61.9}       & 88.0       & 61.1      & \textbf{88.2}     \\
classroom-2             & 42.2       & 71.9      & 43.4       & 74.8       & \textbf{44.9}      & \textbf{75.4}     & 51.6       & 81.3      & 54.6       & \textbf{82.7}       & \textbf{55.9}      & 82.5     & 62.2       & 95.7      & \textbf{68.9}       & 93.2       & 68.3      & \textbf{95.3}     \\ \midrule
Average                & 41.1       & 70.3      & 41.8       & 72.6       & \textbf{46.9}      & \textbf{76.4}     & 47.2       & 76.6      & 48.2       & 77.9       & \textbf{53.8}      & \textbf{81.5}     & 52.2       & 81.3      & 52.7       & 83.0       & \textbf{58.3}      & \textbf{85.5}  \\
\bottomrule
\end{tabular}
\caption{Effectiveness of using region-specific semantic features, context-aware semantic features, and both, when testing on other settings with at least 30\%, 70\%, and 90\% pixel correspondence, respectively.}

\label{tab:other correspondence}
\end{table*}

\noindent\textbf{Selection of Features} We analyze the quality of the learned region-specific features and context-aware features in our model. Particularly, for each scene, 100 samples are randomly selected for the statistical analysis. For each query, we compute the IoU of rendered region-specific and rendered context-aware features, respectively. The one with a higher IoU is considered as the ``ground truth" correct selection of feature type for the current query over the sample. We count the frequency of being the better one based on the ``ground truth" for each type of feature. We can see that each type of features has their advantages and stands out with large opportunities. In addition, we assess how often the feature selection strategy hits the better-performing feature, and Tab.~\ref{tab:accuracy of strategy on rendered feature} shows that the average accuracy across all the scenes reaches 83.7\%, indicating the effectiveness of this strategy. Note that we focus only on cases where the IoU difference between the two types of features exceeds 5\% to exclude the cases where the two types of features have similar performance. The accuracy is different from that evaluated on the extracted features as shown in Tab.~\ref{tab:accuracy of strategy}. That is because the learned semantic features are already very different from those supervision features. For example, the supervision features suffer from view inconsistency while the learned features have delimitated this problem.    

\begin{table}
    \centering
    \small
\setlength{\tabcolsep}{1mm}
    \resizebox{0.42\textwidth}{!}{
    \begin{tabular}{c|c|c|c}
    \toprule
         Scene & Region Pro. & Context Pro. & Accuracy\\ \midrule
         office & 43.4  & 56.6 & 76.5\\
         classroom-1 & 50.9  & 49.1 & 83.0 \\
         utility room & 54.4  & 45.6 & 86.5 \\
         workshop & 63.8 & 36.2 & 92.1\\
         printing room & 45.6  & 54.4 & 84.8\\
         kitchen & 55.0 & 45.0 & 81.1 \\
         meeting room & 22.1 & 77.9 & 85.7\\
         classroom-2 & 26.5 & 73.5 & 80.2 \\ \midrule
         Average & 45.2 & 54.8 & 83.7 \\
    \bottomrule
    \end{tabular}
    }
     \caption{Effectiveness of region-specific features and context-aware features (measured by the probability of being better), and the feature selection strategy, evaluated on the \ourss \emph{rendered} 2D semantic features.}
    \label{tab:accuracy of strategy on rendered feature}
\end{table}

\begin{figure}[!h]
\centering
\includegraphics[width=\linewidth]{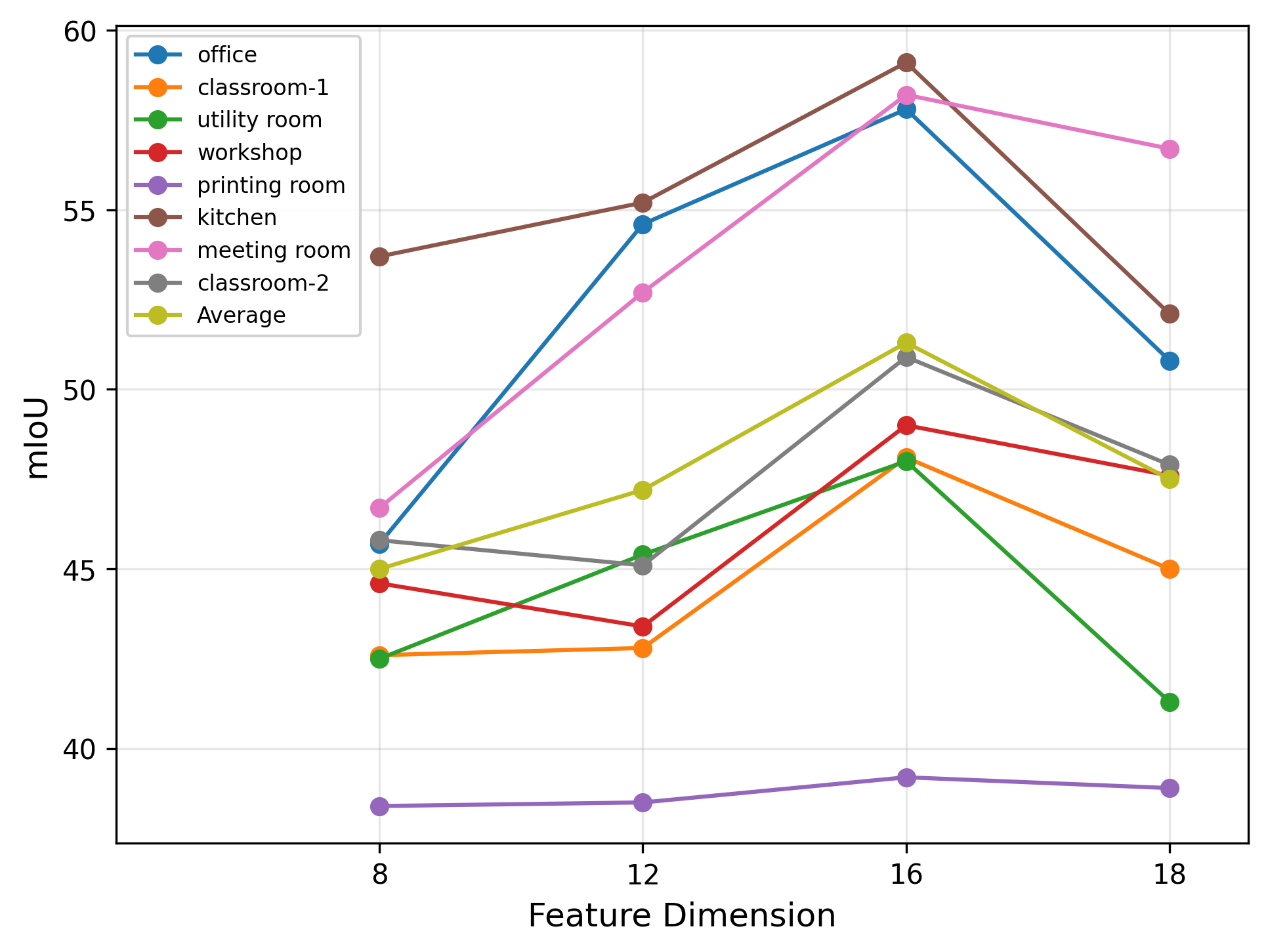} \\
\vspace{-1mm}
\caption{Segmentation performance (mIoU) on different feature dimensions.}
\vspace{-4mm}
\label{fig:mIoU_feature_dim}
\end{figure}
\begin{figure}[!h]
\centering
\includegraphics[width=\linewidth]{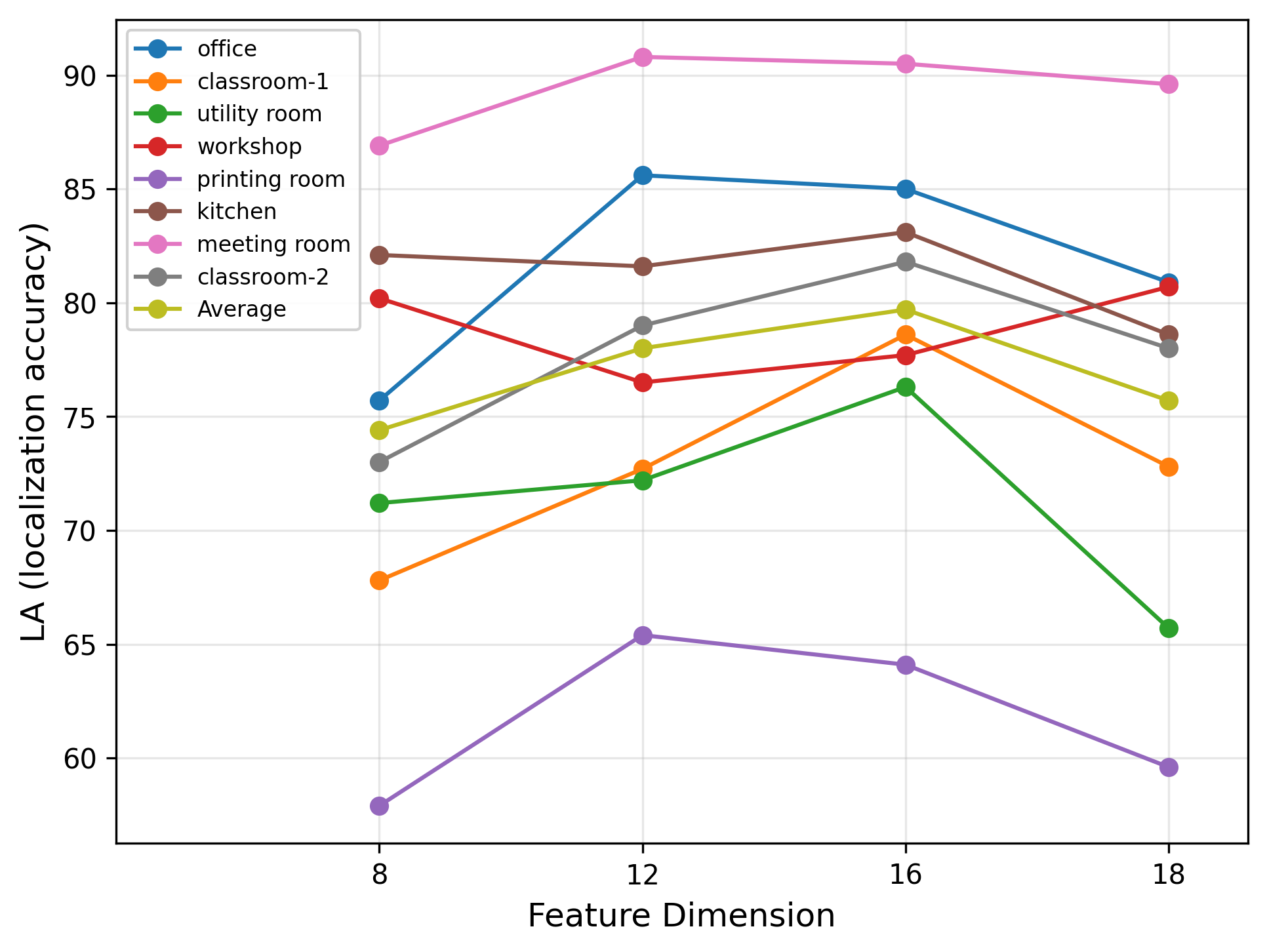} \\
\vspace{-1mm}
\caption{Localization accuracy on different feature dimensions.}
\vspace{-4mm}
\label{fig:la_feature_dim}
\end{figure}

\noindent\textbf{Ablation on Feature Selection Strategies} We choose the region-specific or context-aware features that produce a higher regularized relevancy score as the selected features. Tab.~\ref{tab:querying strategies} compares performance using different strategies. Strategy ``Mean'' involves averaging the two types of rendered features for querying. Besides, we compare with the ``Upper Bound" strategy: after obtaining region-specific and context-aware features, for a query to a rendered image, we calculate the IoU with the ground truth segmentation, respectively, and choose the one with the higher IoU (the same process applies to localization accuracy). Compared with ``Mean'', our strategy achieves a higher performance, approximating the ``Upper Bound''. Note that for each scene, 100 samples are randomly selected for the statistical analysis.

\begin{table}[!h]
\centering
\small
\setlength{\tabcolsep}{1mm}
\resizebox{0.42\textwidth}{!}{
\begin{tabular}{c|cc|cc|cc}
\toprule
\multirow{2}{*}{Scene} & \multicolumn{2}{c|}{Mean} & \multicolumn{2}{c|}{Ours} & \multicolumn{2}{c}{Upper Bound} \\ \cmidrule{2-7}
                   & mIoU$\uparrow$ & LA$\uparrow$          & mIoU$\uparrow$ & LA$\uparrow$              & mIoU$\uparrow$ & LA$\uparrow$        \\  \midrule
office             & 57.7 & 85.6      & 57.8 & 84.9       & 60.2 & 86.3   \\  
classroom-1        & 47.9 & 78.3      & 48.2 & 78.6       & 49.6 & 79.4   \\ 
utility room       & 38.8 & 70.5      & 47.9 & 76.3       & 49.3 & 77.5    \\ 
workshop           & 47.1 & 79.5      & 48.9 & 77.7       & 49.9 & 80.4    \\ 
printing room      & 36.6 & 61.5      & 39.2 & 64.1       & 40.7 & 68.6   \\ 
kitchen            & 51.7 & 80.3      & 59.1 & 83.1       & 60.7 & 85.4    \\ 
meeting room       & 58.1 & 90.9      & 58.2 & 90.8       & 59.9 & 92.0   \\ 
classroom-2        & 50.1 & 81.7      & 50.9 & 81.9       & 53.0 & 82.8  \\ \midrule
Average            & 48.5 & 78.5      & 51.3 & 79.7       & 52.9 & 81.6  \\
\bottomrule
\end{tabular}
}
\setlength{\tabcolsep}{6pt}
\caption{Ablation study on feature selection strategies for text querying. For each scene, 100 samples are randomly selected for statistical analysis.}
\label{tab:querying strategies}
\end{table}

\noindent\textbf{Ablation on Feature Dimension $K$} In addition to the default feature dimension $K$ set to 16, we set $K$ to 8, 12, and 18 to study its impact. Fig.~\ref{fig:mIoU_feature_dim} and Fig.~\ref{fig:la_feature_dim} show the segmentation performance (mIoU) and localization accuracy on different feature dimensions, respectively. We observe that our method's performance first improves with the increase of feature dimensions and achieves the best average performance when $K$ is 16. The performance declines at higher dimensions, which we guess this is due to the increased difficulty in model optimization with higher dimensionality.

\begin{figure}[!h]
\centering
\includegraphics[width=\linewidth]{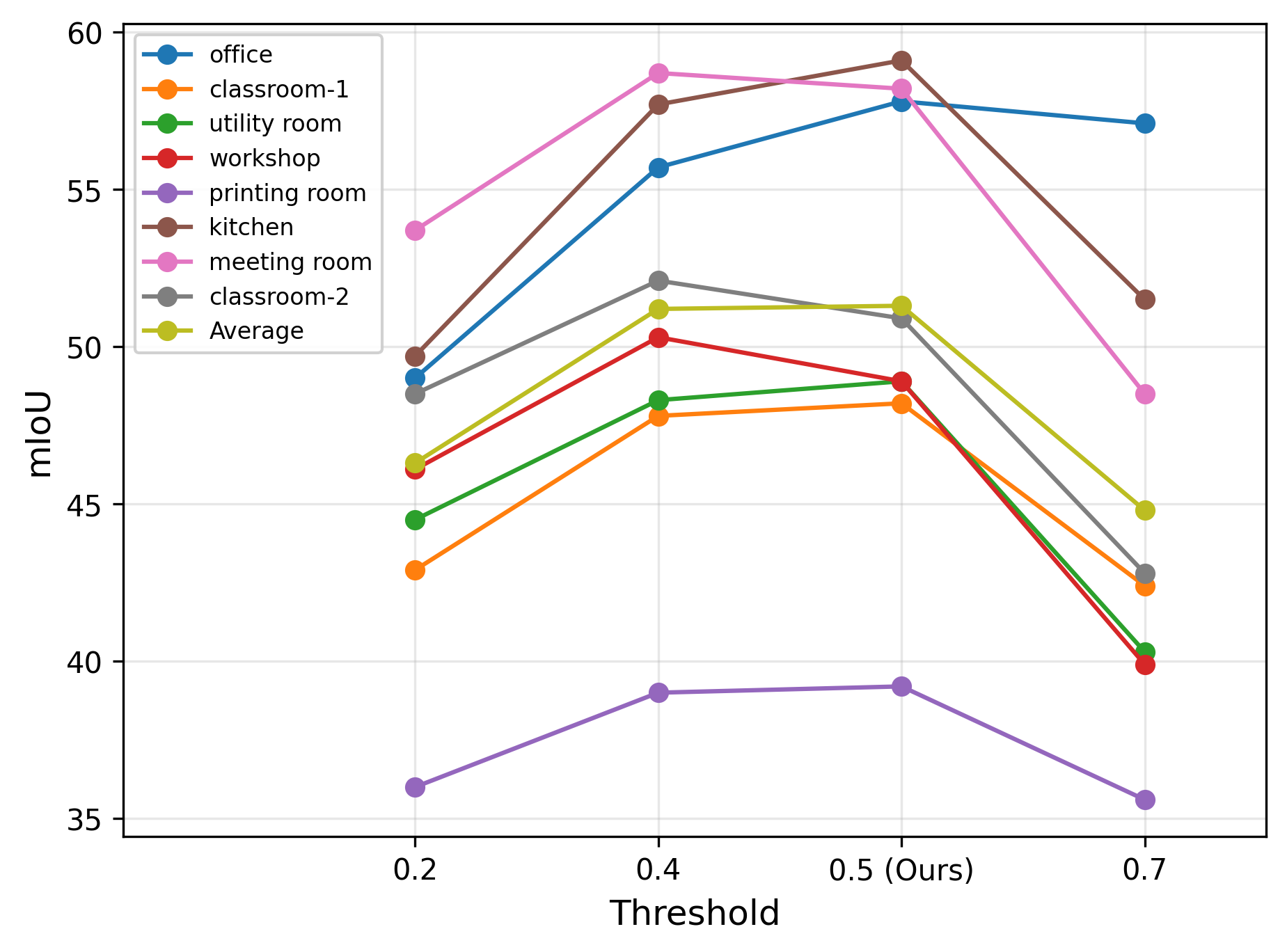} \\
\vspace{-1mm}
\caption{Segmentation performance (mIoU) on different thresholds of the regularized relevance score.}
\vspace{-4mm}
\label{fig:mIoU_thres}
\end{figure}

\begin{figure}[!h]
\centering
\includegraphics[width=\linewidth]{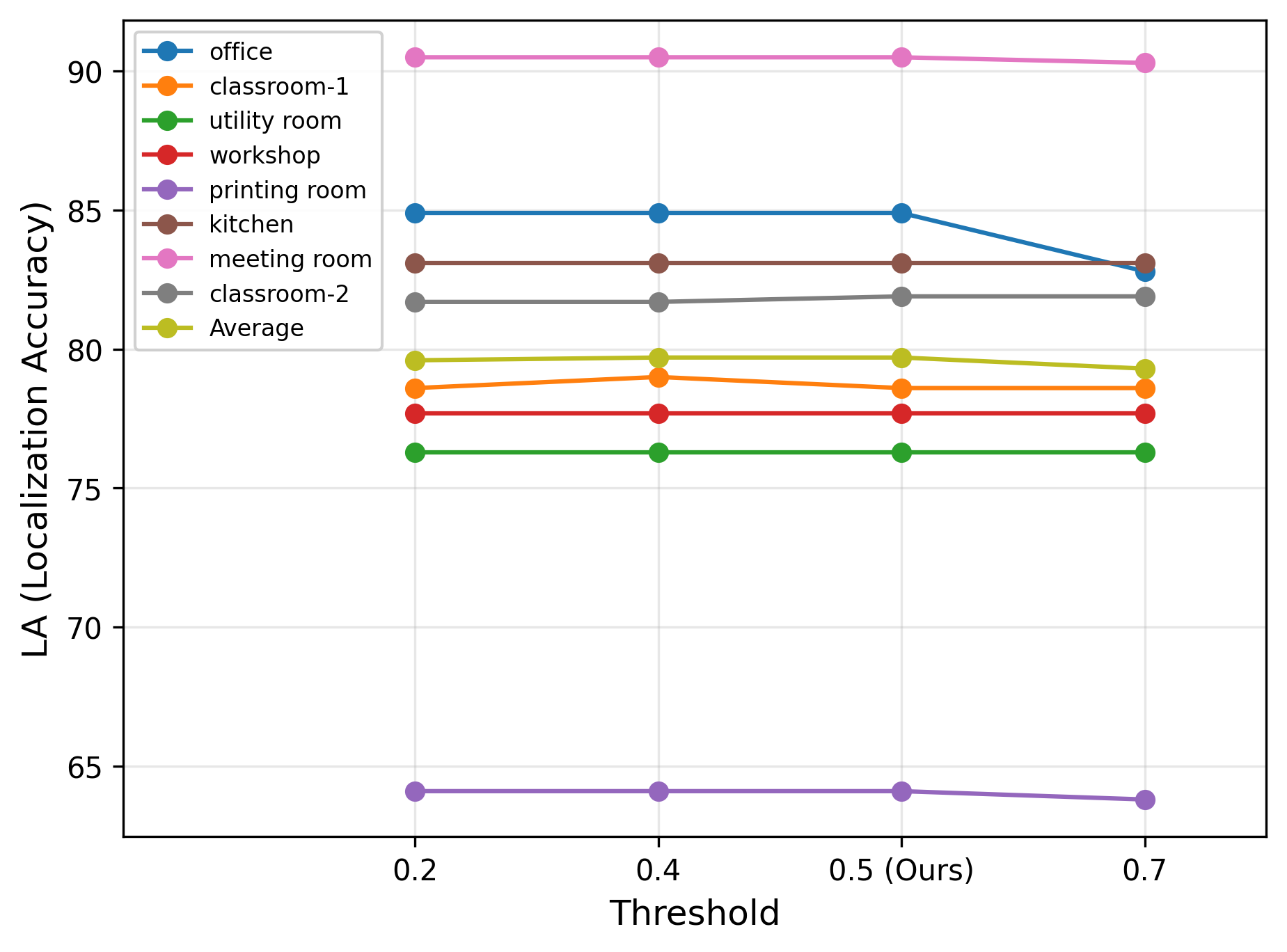} \\
\vspace{-1mm}
\caption{Localization accuracy on different thresholds of the regularized relevance score.}
\vspace{-4mm}
\label{fig:la_thres}
\end{figure}

\noindent\textbf{Ablation on Threshold of Relevancy Score} During open-vocabulary querying, a fixed empirical threshold is manually set to select regions. For each position, when the regularized relevancy score is higher than the threshold, it is identified as the retrieved location. In our experiment, we set the threshold to 0.5 by default. Further analyses with thresholds of 0.2, 0.4, and 0.7 are shown in Fig.~\ref{fig:mIoU_thres} and Fig.~~\ref{fig:la_thres} measured by segmentation performance and localization accuracy, respectively. We can see that the threshold too high or too low negatively influences the performance, whereas thresholds in the range of 0.4 to 0.5 yield better results. 


\begin{table*}[t]
\centering
\normalsize
\small
\setlength{\tabcolsep}{0.45mm}
\resizebox{0.85\textwidth}{!}{ 
\begin{tabular}{c|c|c|c|c|c|c|c|c}
\toprule
\multirow{2}{*}{Method} 
                        & office & classroom-1 & utility room & workshop  & printing room & kitchen & meeting room & classroom-2 \\ \cmidrule{2-9}
                        & PSNR$\uparrow$  & PSNR$\uparrow$    & PSNR$\uparrow$  & PSNR$\uparrow$  & PSNR$\uparrow$  & PSNR$\uparrow$  & PSNR$\uparrow$  & PSNR$\uparrow$ \\ \midrule
LangSplat               & 33.1 & 28.6 & 31.8 & 29.9 & 27.0 & 33.5 & 30.2 & 32.7 \\ \midrule
LangSplat~(using loss masks) & 32.9  & 30.3 & 32.3 & 30.0 & 26.8  & 33.5  & 33.9  & 32.8 \\ \midrule
\ours~(Ours)            & 21.5 & 14.1 & 22.7 & 15.5 & 22.7 & 22.1 & 19.6 & 24.6 \\
\bottomrule
\end{tabular}
}
\caption{RGB quality comparison between LangSplat and our \ourss in terms of PSNR (dB).}
\label{tab:RGB Quality Comparison Results}
\end{table*}

\section{RGB Quality Comparison with LangSplat}

We compare the view synthesis quality of the per-scene optimization-based method LangSplat and the generalizable method. Note our \ourss inherits the view synthesis capability of Splatt3R \cite{smart2024splatt3r} while enabling the capability of semantic understanding. Tab.~\ref{tab:RGB Quality Comparison Results} reports the RGB quality of the synthesized views for all eight scenes as listed in Tab.~\ref{tab:Comparison Results}. Specifically, we compute the PSNR between the rendered images and the ground truth images. Following Splatt3R, the PSNR metric is calculated after applying loss masks in \ours. For LangSplat, we report the PSNR after applying the same loss masks, and the PSNR without applying loss masks. As discussed in Sec.~\ref{sec:discussion} and the results shown in Tab.~\ref{tab:RGB Quality Comparison Results}, 
LangSplat achieves a much higher reconstruction quality due to its costly per-scene optimization under the dense view supervision of RGB images. Even with the per-scene optimization under the dense view semantic supervision, LangSplat is inferior to our \ourss (see Tab.~\ref{tab:Comparison Results}). The reason is analyzed in Sec.~\ref{sec:discussion}, where our generalizable model is robust to the interference of noise in semantic features. 

\section{Failure Cases}

\begin{figure}
    \centering
    \includegraphics[width=\linewidth]{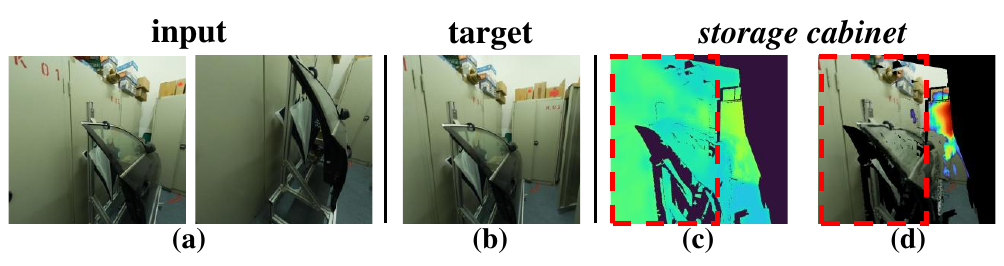}
    \caption{Failure case 1: the ``storage cabinet" region is only partially identified. The region highlighted by the red box actually has a high relevance score but it is still lower than the given threshold and thus is not identified. }
    \label{fig:fail thres}
\end{figure}

\begin{figure}
    \centering
    \includegraphics[width=\linewidth]{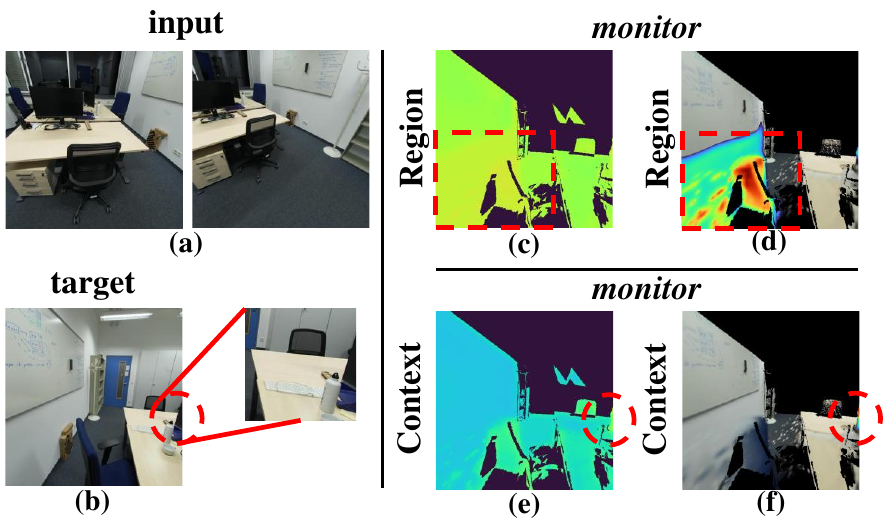}
    \caption{Failure case 2: the context features identify the monitor region correctly but region-specific features are selected for the query of ``monitor" (\emph{i.e.}, the feature selection strategy fails here).}
    \label{fig:fail strategy}
\end{figure}

\begin{table*}[t]
    \centering
    \small
    \setlength{\tabcolsep}{1mm}
    \resizebox{0.95\textwidth}{!}{ 
    \begin{tabular}{c|c|l|c|c}
    \toprule
        \multirow{2}{*}{Scene} & \multirow{2}{*}{Scene IDs} & \multicolumn{1}{c|}{\multirow{2}{*}{Queries}} & \multicolumn{2}{c}{Image IDs} \\ \cmidrule{4-5}
        & & & Input Images & Target Images \\ \midrule
        office    & 7b6477cb95 & blinds, ceiling, door, floor, monitor, office chair, storage cabinet, table, wall, whiteboard & 3679, 3738 & 3688 \\
        classroom-1    & bde1e479ad & ceiling, door, floor, table, wall, whiteboard & 2303, 2369  & 2212 \\
        utility room   & e398684d27 & box, ceiling, door, floor, storage cabinet, wall & 9837, 9987 & 9839 \\
        workshop       & c5439f4607 & cabinet, ceiling, door, floor, office chair, wall & 5760, 6230  & 5750 \\
        printing room  & 825d228aec & box, ceiling, door, floor, shelf, table, wall & 2313, 2320  & 2443 \\
        kitchen        & 40aec5fffa & door, floor, kitchen cabinet, refrigerator, wall & 9693, 9771  & 9696 \\
        meeting room   & fb5a96b1a2 & ceiling, chair, floor, table, wall, whiteboard & 2981, 2998  & 2789 \\
        classroom-2    & c4c04e6d6c & ceiling, chair, door, floor, table, wall & 3166, 3349 & 3164 \\
    \bottomrule
    \end{tabular}
    }
    \caption{Details of scenes related to the scene IDs,  queries, and image IDs from the ScanNet++ dataset.}
    \label{tab:scene_details}
\end{table*}

We analyze the failure cases to understand the limitations of our method. 1) Sometimes, our \ourss fails to segment all regions that correspond to the query text since the relevancy score is lower than the threshold even though the score is not very low. As the example shown in Fig.~\ref{fig:fail thres} (c), the ``storage cabinet" region as highlighted by the red box exhibits an obvious response to the query but is not high enough. As a result, with a fixed threshold, it is filtered out, leaving only regions with higher responses as shown in Fig.~\ref{fig:fail thres} (d). 2) Our feature selection strategy prioritizes the feature with a higher response to the input text. This strategy succeeds at a high probability but sometimes fails. By comparing Fig.~\ref{fig:fail strategy} (c) and (e) which represent the responses of the region-specific features and the responses of the context-aware features, we can see that the region-specific feature exhibits a stronger response. However, the selected regions are focused on the ``wall'' and ``office chair'', which deviate significantly from the query. In contrast, although the context-aware feature shows a slightly weaker response, it accurately localizes the ``monitor''. We think that exploring more effective strategies for selecting or integrating different features would further improve the performance. Moreover, an adaptive determination of thresholds would be helpful to reduce missed detections.   

\section{Details of Scene Selection}
Tab.~\ref{tab:scene_details} presents the details of the selected eight representative scenes from ScanNet++, including the scene IDs in the dataset, the typical categories chosen as query texts for each scene, and the image IDs for the two input/reference images and one target image (used in the per-scene comparison experiments with LangSplat).

\end{document}